\documentclass[journal]{IEEEtran}
\usepackage{authblk}
\usepackage{graphicx}
\usepackage{csquotes}
\usepackage{xurl}
\usepackage{tabularx}
\usepackage{float}
\usepackage{hyperref}
\usepackage{caption}
\usepackage{textcomp}
\usepackage{siunitx}

\ifCLASSINFOpdf
\else
\fi
\begin{document}
%
\title{Back Translation Survey for Improving Text Augmentation}
%
%
%

\author[1]{\small Matthew Ciolino}
\author[1]{\small David Noever}
\author[1, 2]{\small Josh Kalin}

\affil[1]{\footnotesize PeopleTec, Inc, 4901 Corporate Dr NW, Huntsville, AL 35805, USA}
\affil[2]{\footnotesize Department of Computer Science and Software Engineering, Auburn University, Auburn, AL, USA}


%
%

\markboth{PREPRINT}%
{Shell \MakeLowercase{\textit{et al.}}: Bare Demo of IEEEtran.cls for IEEE Journals}
%



\maketitle

\begin{abstract}
Natural Language Processing (NLP) relies heavily on training data. Transformers, as they have gotten bigger, have required massive amounts of training data. To satisfy this requirement, text augmentation should be looked at as a way to expand your current dataset and to generalize your models. One text augmentation we will look at is translation augmentation. We take an English sentence and translate it to another language before translating it back to English. In this paper, we look at the effect of 108 different language back translations on various metrics and text embeddings.
\end{abstract}

\begin{IEEEkeywords}
Embeddings, Translations
\end{IEEEkeywords}

%
\IEEEpeerreviewmaketitle

\section{Introduction}
Training machine learning models have always required an increasing amount of data for the increasing size of the architecture. This phenomenon has been exemplified in the NLP community with the recent explosion in use of the transformer \cite{vaswani2017attention}. Public datasets reflect this need for larger and larger data requirements in the recently released the 800GB Pile dataset \cite{gao2020pile}. In the absence of readily available data, an unsupervised way of text augmentation is needed.

Back translation is the unsupervised process of translating text from English to another language and then back to English. This text augmentation technique allows a variety of outputs for any input. A strategy might be to translate from English to many other languages and then back to English to create the most broad understanding of the input text. In this scenario, we envision that a system of back translations can provide transformers with the generalized data that they need to train larger and larger models. 

To employ a complete text augmentation training strategy we need to understand the full effect of back translation. In this paper, we will investigate how each language's back translation effects various NLP metrics. We hope to decipher how and why some languages might be a better candidate for back translating to over any other language. We employ Google Translate's current Neural Machine Translator (NMT) \cite{wu2016google} to receive over 108 languages translations. 

\subsection{Background}
Using text augmentation is a unsupervised way to expand your text training data. There are various text augmentations \cite{chaudhary2020nlpaugment}, for example the nlpaug package \cite{ma2019nlpaug} summarizes them in 5 categories: insert, substitute, swap, delete or crop. Some examples from the nlpaug pacakge include: RandomWordAug (Apply augmentation randomly), AntonymAug (Substitute opposite meaning word according to WordNet antonym), SplitAug (Split one word to two words randomly), WordEmbsAug (Leverage word2vec, GloVe or fasttext embeddings to apply augmentation).

Back translation is a version of the substitute augmentation. We take the imperfect system of translation in an attempt to increases generalizability of text models. This text augmentation has shown great performance for various tasks including text classification \cite{wei2019eda}, machine translation \cite{sennrich2015improving} \cite{edunov2018understanding}, and even in low data environments \cite{fadaee2017data}. While back translation has been show to improve various NLP tasks, in some situations is has been shown to provide marginal, if any, results in modern, large transformer \cite{longpre2020effective}.

In addition, to showing various NLP metrics for all 108 Google Translate supported languages, we also update the BLEU metrics for the Google's current NMT model. Previous scores generated by Aiken in 2019 can be found here \cite{aiken2019updated}. Of note is that all languages besides Latin use Google's NMT while Latin uses Google's Phrase-Based Machine Translation (PBMT).

\subsection{Reproducible}
To allow reproducible for the following experiment a Google Collaboratory \cite{google}. You can replicate the experiment in the following notebook. This colab includes all training data and models used. \smallskip

\begin{quote}
\textbf{Reproducible Google Colaboratory} \\
\url{https://colab.research.google.com/drive/1XpdkDrNruJ5TDZlwtdSRgZlkDRjcUM7d?usp=sharing}
\end{quote}

\section{Experiment}
\begin{figure}[ht]
    \centering
    \includegraphics[width=.485\textwidth]{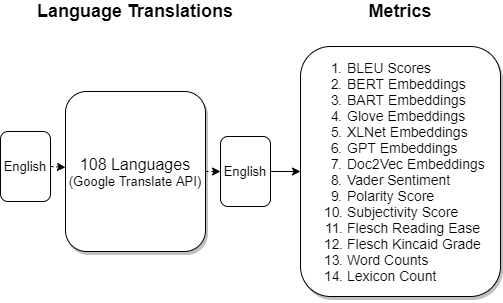}
    \caption{Experimental flow diagram showing the languages used for translations followed by the metrics used to analysis the differences}
    \label{fig:exp}
\end{figure}

To investigate the effect of different language translations on various text metrics we run a series of test as outlined in [Figure \ref{fig:exp}]. We took 1000 random English tweets from the Sentiment-140 \cite{go2009twitter} dataset, and run them through the Google Translate API \cite{googletranslate}. We translate to all 108 languages supported by Google Translate and then translated them back to English. We then analysed the differences using various text metrics: Bilingual Evaluation Understudy Score (BLEU) \cite{papineni2002bleu}, BERT embedding distance \cite{devlin2018bert}, BART embedding distance \cite{lewis2019bart}, GPT embedding distance \cite{radford2018improving}, XLNet embedding distance \cite{yang2019xlnet}, GloVe embedding distance \cite{pennington2014glove}, Doc2Vec embedding distance \cite{le2014distributed}, NLTK Vader \cite{hutto2014vader}, Textblob Polarity and Subjectivity \cite{loria2018textblob}, Flair \cite{akbik2018coling}, and common Text Statistics (Flesch-Kincaid Grade, Flesch-Reading Ease) \cite{flesch1943marks}. The following sections go into detail about each of these NLP metrics.

\subsection{BLEU Scores}
BLEU is a metric for evaluating a generated sentence to a reference sentence. Scores a translation on a scale of 0 to 1, in an attempt to measure the adequacy and fluency of the machine translation output. Scored by n-grams, BLEU-1 to BLEU-4. We use NLTK's BLEU \cite{nltkbleu} score method which can weight each of the n-grams scores independently.

\subsection{BERT Embeddings}
Bidirectional Encoder Representations from Transformer (BERT) is an encoder encoder transformer architecture that trains by predicting masked words. By masking 15\% of the words, BERT can also predict the position of words in a sentence. This allows BERT to be a general language model which can predict word embeddings alongside their positional embeddings. BERT was originally pre-trained on the English Wikipedia and Brown Corpus. Our implementation used bert-base-uncased \cite{hugging_face}, 12-layer, 768-hidden, 12-heads, 110M parameters.

\subsection{BART Embeddings}
BART is a transformer that learns to train by generalizing the masking technique used by BERT to a random shuffle of the ordering of a sentence and a mask that spans many words. This unsupervised technique learns to map corrupted parts of a document and therefore performed SOTA on discriminative and text generation tasks in 2020. We use BART-Base \cite{facebook/bart-base}, 12-layer, 768-hidden, 16-heads, 139M parameters, for our analysis.

\subsection{GPT Embeddings}
GPT uses a decoder only transformer structure with masked self-attention to train the language mode. Originally published in 2018, GPT, once fine tuned for a task, was SOTA in many language tasks. This unsupervised pre-training transform architecture set the standard for the large transformer models to follow. We use open-gpt, GPT 1,  \cite{gpt_huggingface}, 12-layer, 768-hidden, 12-heads, 110M parameters.

\subsection{XLNet Embeddings}
XLNet is a generalized autoregressive pretraining method that enables learning bidirectional contexts by maximizing the expected likelihood over all permutations of the factorization order. When released in 2020, XLNet outperformed BERT in many language tasks. We use xlnet-base-cased \cite{xlnet_huggingface}, 12-layer, 768-hidden, 12-heads, 110M parameters.

\subsection{GloVe Embeddings}
Global Vectors for Word Representation, GloVe, is an unsupervised learning algorithm for obtaining vector representations for words. Training is performed on aggregated global word-word co-occurrence statistics from a corpus, and the resulting representations showcase interesting linear substructures of the word vector space \cite{pennington}. GloVe was originally trained on 800 MB of text in Wikipedia 2014 and Gigaword 5. The model outputs a 100d vector.

\subsection{Doc2Vec Embeddings}
Doc2Vec in an extension of the Word2Vec \cite{mikolov2013efficient} architecture which uses either a skip-gram or continuous bag of words method. In addition to learning word vectors, Doc2Vec, also learns a paragraph vector which allows it to learn from documents of any length and format. We use the Associated Press News Skip-gram \cite{jhlau20168doc2vec} \cite{lau2016empirical} (0.6GB) which was trained on Wikipedia and AP News.

\subsection{Vader Compound Score}
Vader is a simple ruled based sentiment analysis tool original made for real-time social media. Vader is constructed from a generalizable,  valence-based, human-curated gold standard sentiment lexicon.
They rule based approach was developed to be sensitive to both the polarity and the intensity of sentiments. To handle English idioms, a special rule set was constructed (the shit:+3, the bomb:+3, bad ass:+1.5, yeah right:-2, cut the mustard:+2, kiss of death:+1.5, hand to mouth:-2).

\subsection{Textblob Polarity/Subjectivity Score}
Textblob employs a lexicon of words \cite{Sloria} (with cornetto-synset-id and wordnet-id) and their part of speech, definition, polarity, subjectivity, and intensity. For example, 
\begin{displayquote}
    $<$ word form="great", pos="JJ", sense="very good", polarity="1.0", subjectivity="1.0", intensity="1.0",  confidence="0.9" $>$
\end{displayquote}
Textblob uses the lexicon from the deprecated Pattern Library \cite{Pattern.web} (found via the WayBack Machine) which contains 2917 entries. We use both the Polarity score and the Subjectivity score in our analysis.

\subsection{Text Statistics}
Text statistics include Flesch-Reading Ease, Flesch-Kincaid Grade, lexicon count (number of words) and sentence count (number of sentences). Flesch metrics. Both Flesch-Reading Ease (Range 0-100) and Flesch-Kincaid Grade (Range 0-18) \cite{readable_2020} are  metrics which uses total words, total sentences, and total syllables to calculate readability. The Flesch Reading Ease score is between 1 and 100, and the Flesch Kincaid Grade Level reflects the US education system.

\section{Evaluation}

We find the following 14 comparison  metrics: BLEU, BERT Embeddings, BERT Embeddings, GPT Embeddings, XLNet Embeddings, Glove Embeddings, Doc2Vec Embeddings, Flair, Polarity, Subjectivity, Vader, Flesch-Kincaid Grade, Flesch Reading Ease, lexicon counts, and word counts for 108 languages. A summarization of all charts is located in the appendix alongside example back translations [Table \ref{tab:tweet}, Table \ref{tab:metrics1}, Table \ref{tab:metrics2}].

\begin{table*}[ht]
\centering
\begin{tabular}{||l|c|c|c|c|l|c|c|c|c||}
\hline
\multicolumn{1}{||c|}{\textbf{Language}} &
  \textbf{BLEU-1} &
  \textbf{BLEU-2} &
  \textbf{BLEU-3} &
  \textbf{BLEU-4} &
  \multicolumn{1}{c|}{\textbf{Language}} &
  \textbf{BLEU-1} &
  \textbf{BLEU-2} &
  \textbf{BLEU-3} &
  \textbf{BLEU-4} \\ \hline
Afrikaans             & 0.6619 & 0.4684 & 0.2483 & 0.2483 & Lithuanian                     & 0.5966 & 0.3759 & 0.1793 & 0.1793 \\ \hline
Albanian              & 0.6637 & 0.4643 & 0.2565 & 0.2565 & Luxembourgish                  & 0.6229 & 0.4044 & 0.1845 & 0.1845 \\ \hline
Amharic               & 0.4862 & 0.2171 & 0.0631 & 0.0631 & Macedonian                     & 0.6411 & 0.4356 & 0.2239 & 0.2239 \\ \hline
Arabic                & 0.5174 & 0.3013 & 0.1215 & 0.1215 & Malagasy                       & 0.5058 & 0.2799 & 0.1051 & 0.1051 \\ \hline
Armenian              & 0.6042 & 0.3794 & 0.1635 & 0.1635 & Malay                          & 0.6413 & 0.4255 & 0.2115 & 0.2115 \\ \hline
Azerbaijani           & 0.5265 & 0.2764 & 0.0918 & 0.0918 & Malayalam                      & 0.3872 & 0.1668 & 0.0446 & 0.0446 \\ \hline
Basque                & 0.6104 & 0.3465 & 0.1415 & 0.1415 & Maltese                        & 0.7265 & 0.5309 & 0.3185 & 0.3185 \\ \hline
Belarusian            & 0.5683 & 0.3440 & 0.1443 & 0.1443 & Maori                          & 0.5603 & 0.3192 & 0.1295 & 0.1295 \\ \hline
Bengali               & 0.5269 & 0.2744 & 0.0980 & 0.0980 & Marathi                        & 0.4848 & 0.2289 & 0.0704 & 0.0704 \\ \hline
Bosnian               & 0.6025 & 0.3910 & 0.1842 & 0.1842 & Mongolian                      & 0.4601 & 0.2050 & 0.0502 & 0.0502 \\ \hline
Bulgarian             & 0.5981 & 0.3862 & 0.1775 & 0.1775 & Myanmar (Burmese)              & 0.4885 & 0.2321 & 0.0661 & 0.0661 \\ \hline
Catalan               & 0.5504 & 0.3129 & 0.1292 & 0.1292 & Nepali                         & 0.4926 & 0.2297 & 0.0716 & 0.0716 \\ \hline
Cebuano               & 0.6335 & 0.4292 & 0.2137 & 0.2137 & Norwegian                      & 0.7083 & 0.5470 & 0.3193 & 0.3193 \\ \hline
Chinese (Simplified)  & 0.4817 & 0.2509 & 0.0853 & 0.0853 & Nyanja (Chichewa)              & 0.5212 & 0.2871 & 0.1064 & 0.1064 \\ \hline
Chinese (Traditional) & 0.4830 & 0.2525 & 0.0860 & 0.0860 & Odia (Oriya)                   & 0.5146 & 0.2322 & 0.0625 & 0.0625 \\ \hline
Corsican              & 0.6271 & 0.4044 & 0.1961 & 0.1961 & Pashto                         & 0.4685 & 0.2354 & 0.0659 & 0.0659 \\ \hline
Croatian              & 0.6042 & 0.3959 & 0.1882 & 0.1882 & Persian                        & 0.4977 & 0.2967 & 0.1238 & 0.1238 \\ \hline
Czech                 & 0.6071 & 0.3877 & 0.1819 & 0.1819 & Polish                         & 0.5980 & 0.3653 & 0.1555 & 0.1555 \\ \hline
Danish                & 0.7205 & 0.5678 & 0.3566 & 0.3566 & Portuguese & 0.6570 & 0.4500 & 0.2345 & 0.2345 \\ \hline
Dutch                 & 0.6674 & 0.4725 & 0.2516 & 0.2516 & Punjabi                        & 0.5354 & 0.2910 & 0.1042 & 0.1042 \\ \hline
English               & 1.0000 & 0.9920 & 0.9500 & 0.9500 & Romanian                       & 0.6199 & 0.3973 & 0.1877 & 0.1877 \\ \hline
Esperanto             & 0.6968 & 0.5320 & 0.3157 & 0.3157 & Russian                        & 0.5828 & 0.3611 & 0.1551 & 0.1551 \\ \hline
Estonian              & 0.6044 & 0.3811 & 0.1696 & 0.1696 & Samoan                         & 0.5400 & 0.3133 & 0.1248 & 0.1248 \\ \hline
Finnish               & 0.6156 & 0.4041 & 0.1882 & 0.1882 & Scots Gaelic                   & 0.6349 & 0.4170 & 0.1965 & 0.1965 \\ \hline
French                & 0.6346 & 0.4226 & 0.2005 & 0.2005 & Serbian                        & 0.4944 & 0.2988 & 0.1206 & 0.1206 \\ \hline
Frisian               & 0.7255 & 0.5640 & 0.3481 & 0.3481 & Sesotho                        & 0.5414 & 0.2961 & 0.1114 & 0.1114 \\ \hline
Galician              & 0.5937 & 0.3644 & 0.1627 & 0.1627 & Shona                          & 0.5270 & 0.3064 & 0.1154 & 0.1154 \\ \hline
Georgian              & 0.6055 & 0.3761 & 0.1570 & 0.1570 & Sindhi                         & 0.4755 & 0.2260 & 0.0597 & 0.0597 \\ \hline
German                & 0.6345 & 0.4075 & 0.1875 & 0.1875 & Sinhala (Sinhalese)            & 0.4646 & 0.1980 & 0.0506 & 0.0506 \\ \hline
Greek                 & 0.6374 & 0.4255 & 0.2104 & 0.2104 & Slovak                         & 0.5910 & 0.3779 & 0.1707 & 0.1707 \\ \hline
Gujarati              & 0.5472 & 0.3017 & 0.1144 & 0.1144 & Slovenian                      & 0.5979 & 0.3754 & 0.1636 & 0.1636 \\ \hline
Haitian Creole        & 0.6886 & 0.4966 & 0.2814 & 0.2814 & Somali                         & 0.5905 & 0.3468 & 0.1525 & 0.1525 \\ \hline
Hausa                 & 0.6460 & 0.4540 & 0.2395 & 0.2395 & Spanish                        & 0.6001 & 0.3697 & 0.1671 & 0.1671 \\ \hline
Hawaiian              & 0.4993 & 0.2785 & 0.0978 & 0.0978 & Sundanese                      & 0.6523 & 0.4382 & 0.2261 & 0.2261 \\ \hline
Hebrew                & 0.5618 & 0.3377 & 0.1446 & 0.1446 & Swahili                        & 0.6642 & 0.4616 & 0.2466 & 0.2466 \\ \hline
Hindi                 & 0.5345 & 0.2919 & 0.1050 & 0.1050 & Swedish                        & 0.6728 & 0.4985 & 0.2816 & 0.2816 \\ \hline
Hmong                 & 0.6311 & 0.4414 & 0.2253 & 0.2253 & Tagalog (Filipino)             & 0.7298 & 0.5561 & 0.3349 & 0.3349 \\ \hline
Hungarian             & 0.5593 & 0.3183 & 0.1202 & 0.1202 & Tajik                          & 0.5657 & 0.3265 & 0.1360 & 0.1360 \\ \hline
Icelandic             & 0.6535 & 0.4635 & 0.2459 & 0.2459 & Tamil                          & 0.5012 & 0.2494 & 0.0792 & 0.0792 \\ \hline
Igbo                  & 0.5998 & 0.4059 & 0.1913 & 0.1913 & Tatar                          & 0.4109 & 0.1146 & 0.0122 & 0.0122 \\ \hline
Indonesian            & 0.6353 & 0.4193 & 0.2092 & 0.2092 & Telugu                         & 0.5430 & 0.2867 & 0.1006 & 0.1006 \\ \hline
Irish                 & 0.6893 & 0.4715 & 0.2560 & 0.2560 & Thai                           & 0.4908 & 0.2598 & 0.0955 & 0.0955 \\ \hline
Italian               & 0.6464 & 0.4451 & 0.2332 & 0.2332 & Turkish                        & 0.5029 & 0.2519 & 0.0786 & 0.0786 \\ \hline
Japanese              & 0.4453 & 0.1957 & 0.0541 & 0.0541 & Turkmen                        & 0.4544 & 0.1843 & 0.0441 & 0.0441 \\ \hline
Javanese              & 0.6430 & 0.4247 & 0.2033 & 0.2033 & Ukrainian                      & 0.5636 & 0.3300 & 0.1297 & 0.1297 \\ \hline
Kannada               & 0.5162 & 0.2651 & 0.0914 & 0.0914 & Urdu                           & 0.4734 & 0.2453 & 0.0790 & 0.0790 \\ \hline
Kazakh                & 0.4761 & 0.2144 & 0.0572 & 0.0572 & Uyghur                         & 0.4734 & 0.2012 & 0.0531 & 0.0531 \\ \hline
Khmer                 & 0.5303 & 0.2946 & 0.1142 & 0.1142 & Uzbek                          & 0.4630 & 0.2110 & 0.0559 & 0.0559 \\ \hline
Kinyarwanda           & 0.5368 & 0.2983 & 0.1142 & 0.1142 & Vietnamese                     & 0.6403 & 0.4205 & 0.2098 & 0.2098 \\ \hline
Korean                & 0.4448 & 0.1980 & 0.0578 & 0.0578 & Welsh                          & 0.6812 & 0.4651 & 0.2434 & 0.2434 \\ \hline
Kurdish               & 0.5965 & 0.3650 & 0.1605 & 0.1605 & Xhosa                          & 0.5184 & 0.2804 & 0.1002 & 0.1002 \\ \hline
Kyrgyz                & 0.4033 & 0.1617 & 0.0371 & 0.0371 & Yiddish                        & 0.6307 & 0.5009 & 0.3131 & 0.3131 \\ \hline
Lao                   & 0.5935 & 0.3590 & 0.1626 & 0.1626 & Yoruba                         & 0.6918 & 0.5222 & 0.3166 & 0.3166 \\ \hline
Latin                 & 0.4375 & 0.1882 & 0.0528 & 0.0528 & Zulu                           & 0.6375 & 0.4318 & 0.2238 & 0.2238 \\ \hline
Latvian               & 0.6509 & 0.4510 & 0.2276 & 0.2276 & \multicolumn{5}{l||}{ } \\ \hline
\end{tabular}
\end{table*}

\null\newpage
\null\newpage

\subsection{BLEU Scores}

These scores remearkbly will follow the trend for Google's Reported BLEU score with some variance for the lack of "high quality" reference sentence that BLEU usaslly requires. Instead, we have publicly scraped Tweets which sometimes only include a pronoun (i.e. @user1234) which Google Translate would just return as is.

The top BLEU-1 scores are for Tagalog (BLEU-1=0.7298), Maltese (BLEU-1=0.7265), Frisian (BLEU-1=0.7255) while the top average BLEU scores across all n-grams (25\% weighted for each) are Danish (Weighted BLEU=0.4776), Frisian (Weighted BLEU=0.4719), Tagalog (Weighted BLEU=0.4619). 

On the other side the bottom BLEU-1 Scores are Malayalam (BLEU-1=0.3872),  Kyrgyz (BLEU-1=0.4033), Tatar (BLEU-1=0.4109) while the bottom weighted BLEU score are Tatar (Weighted BLEU=0.0515), Kyrgyz (Weighted BLEU=0.0973), Malayalam (Weighted BLEU=0.1065).

The lower the BLEU score the less likely the reference sentence matches the back translation. 

\begin{figure*}[ht]
    \centering
    \includegraphics[width=\textwidth]{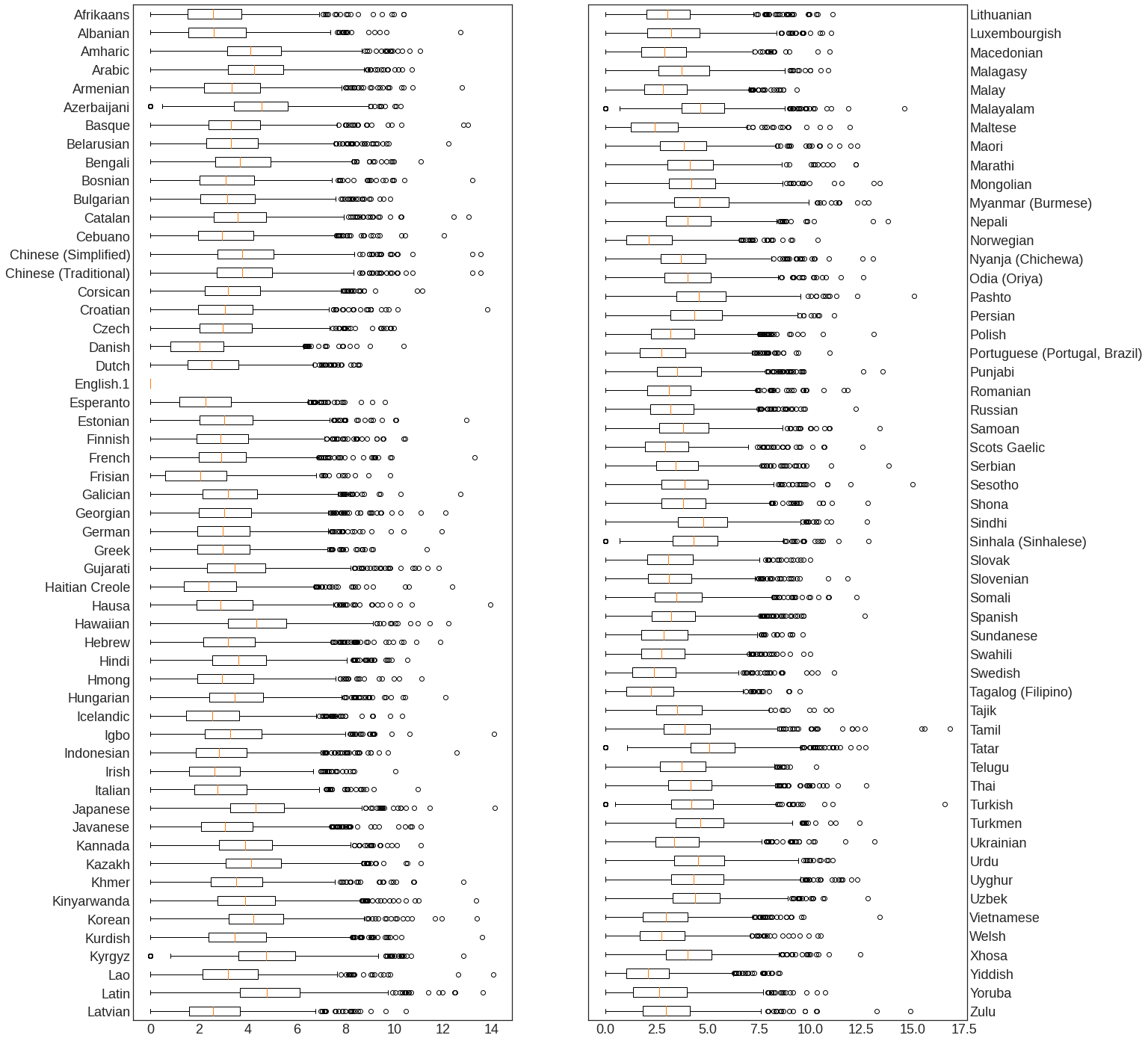}
    \caption{BERT Euclidean Differences}
    \label{fig:1}
\end{figure*}

\null\newpage

\subsection{BERT Embeddings}

Boxplot of absolute distance between BART embeddings. The closer the language is to English the closer the embeddings are to zero. BERT embeddings are of 768 dimensions and the distance is euclidean.

We find closest embeddings are Danish (2.0758 $\pm$ 1.6713), Frisian (2.1105 $\pm$ 1.7137), Yiddish (2.1669 $\pm$ 1.6737). The furthest embeddings from the English reference are Tatar (5.2999 $\pm$ 1.8633), Latin (4.9107 $\pm$ 2.0689), Sindhi (4.8234 $\pm$ 1.9959)

\null\newpage

\begin{figure*}[ht]
    \centering
    \includegraphics[width=\textwidth]{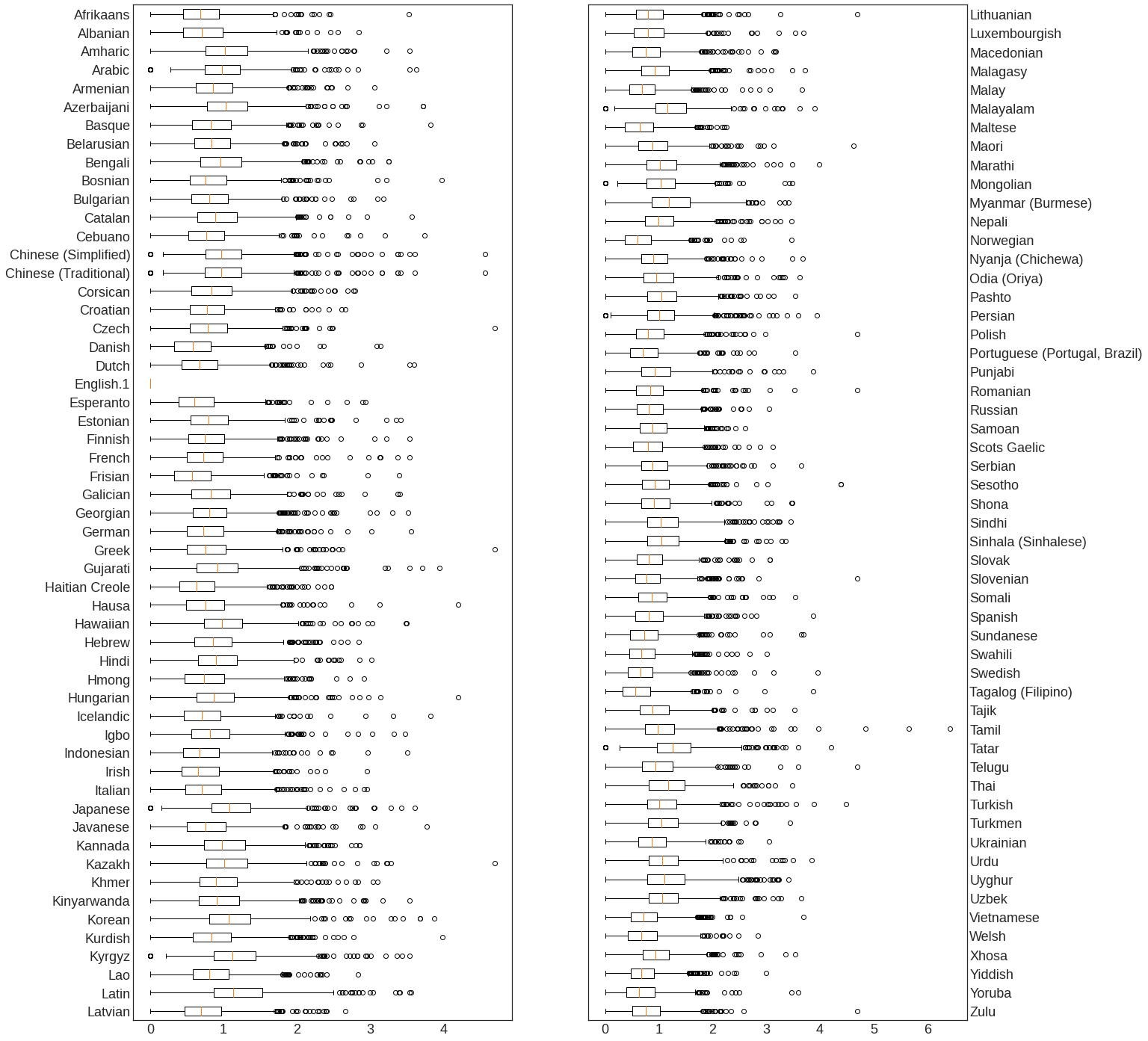}
    \caption{BART Euclidean Differences}
    \label{fig:2}
\end{figure*}

\null\newpage

\subsection{BART Embeddings}

Boxplot of absolute distance between BART embeddings. The closer the language is to English the closer the embeddings are to zero. BART embeddings are of 768 dimensions and the distance is euclidean.

We find closest embeddings are Danish (0.5903 $\pm$ 0.4142), Frisian (0.5954 $\pm$ 0.423), Tagalog (0.5976 $\pm$ 0.4236). The furthest embeddings from the English reference are Tatar (1.309 $\pm$ 0.5083), Myanmar (1.2359 $\pm$ 0.554), Malayalam (1.2327 $\pm$ 0.4586).

\null\newpage

\begin{figure*}[ht]
    \centering
    \includegraphics[width=\textwidth]{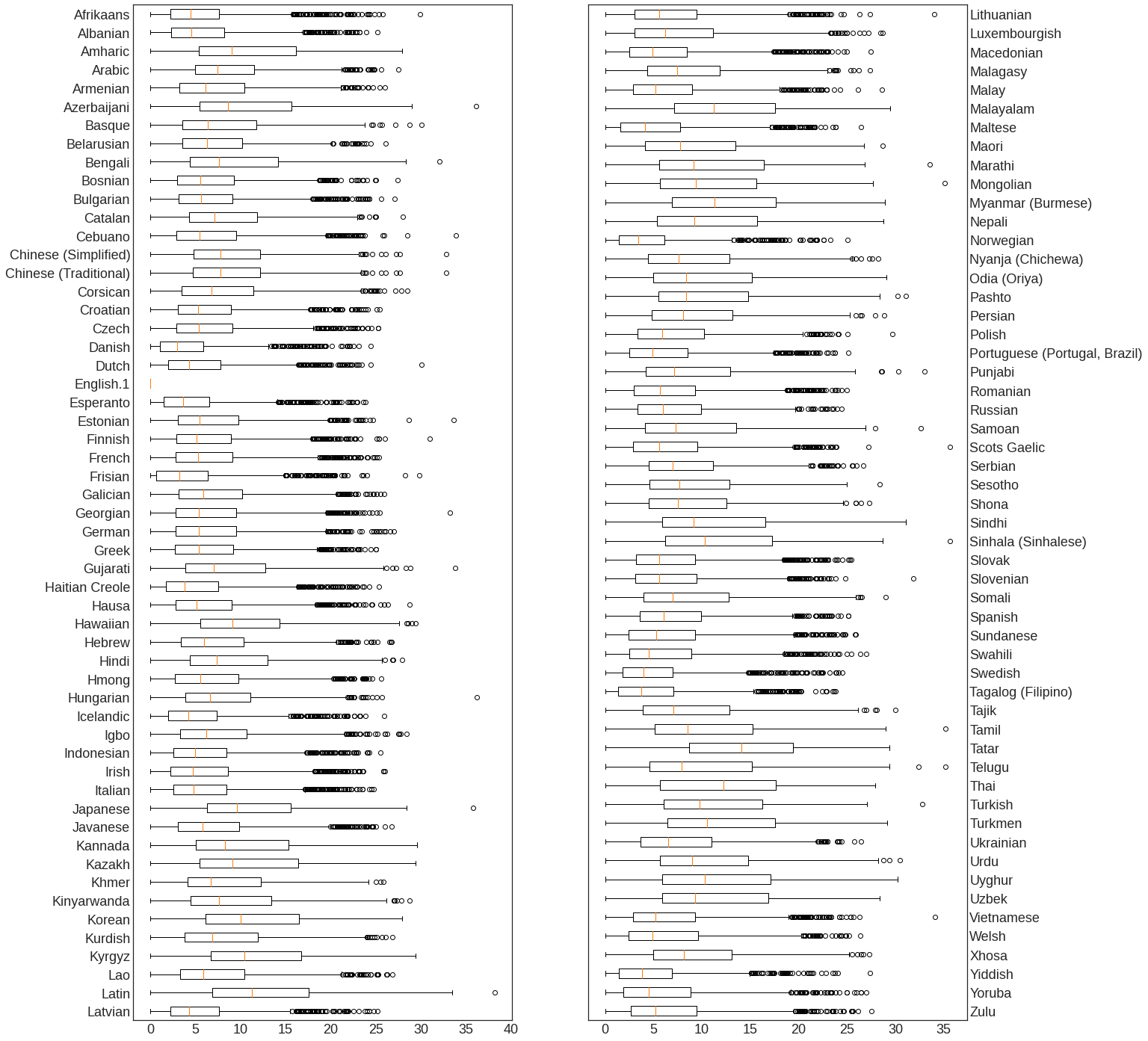}
    \caption{GPT Euclidean Differences}
    \label{fig:2}
\end{figure*}

\null\newpage

\subsection{GPT Embeddings}

Boxplot of absolute distance between GPT embeddings. The closer the language is to English the closer the embeddings are to zero. GPT embeddings are of 768 dimensions and the distance is euclidean.

We find closest embeddings are Danish (4.2798 $\pm$ 4.5897), Norwegian (4.5235 $\pm$ 4.5455), Frisian (4.5661 $\pm$ 5.0261). The furthest embeddings from the English reference are Tatar (13.9721 $\pm$ 6.4386), Malayalam (12.2058 $\pm$ 6.3209), Latin (12.0613 $\pm$ 6.6757).

\null\newpage

\begin{figure*}[ht]
    \centering
    \includegraphics[width=\textwidth]{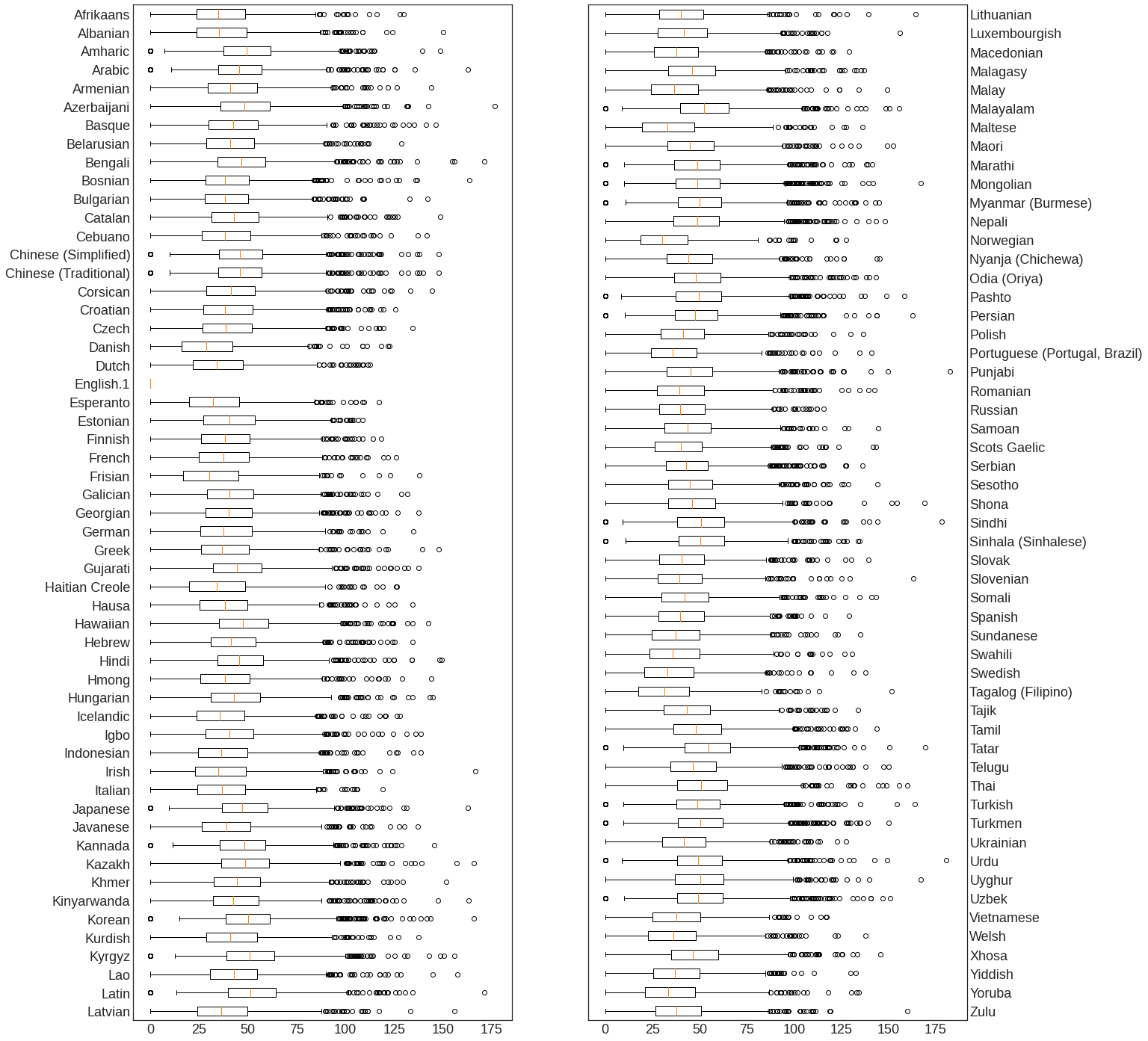}
    \caption{XLnet Euclidean Differences}
    \label{fig:2}
\end{figure*}

\null\newpage

\subsection{XLnet Embeddings}

Boxplot of absolute distance between XLnet embeddings. The closer the language is to English the closer the embeddings are to zero. XLnet embeddings are of 768 dimensions and the distance is euclidean.

We find closest embeddings are Danish (30.1903 $\pm$ 21.0538), Frisian (31.6068 $\pm$ 21.6522), Norwegian (31.6301 $\pm$ 20.7609). The furthest embeddings from the English reference are Tatar (55.9961 $\pm$ 21.5646), Malayalam (54.7046 $\pm$ 21.2162), Kyrgyz (52.7561 $\pm$ 21.0698).

\null\newpage

\begin{figure*}[ht]
    \centering
    \includegraphics[width=\textwidth]{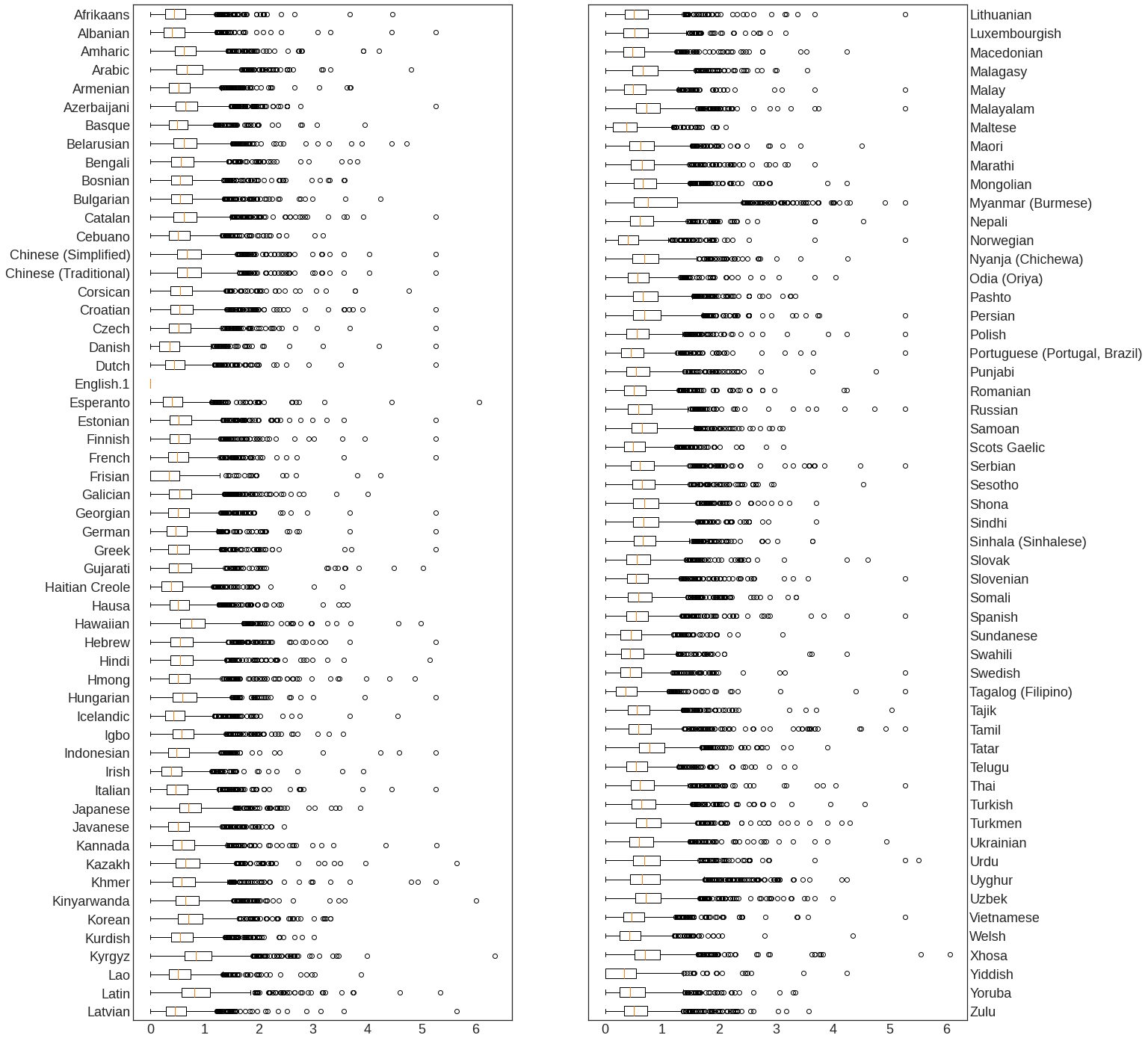}
    \caption{GloVe Pooled Euclidean Differences}
    \label{fig:2}
\end{figure*}

\null\newpage

\subsection{Glove Embeddings}

Boxplot of absolute distance between pooled Glove embeddings. The closer the language is to English the closer the embeddings are to zero. Glove embeddings are of 100 dimensions and the distance is euclidean.

We find closest embeddings are Frisian (0.3772 $\pm$ 0.38), Maltese (0.3895 $\pm$ 0.3353), Yiddish (0.3952 $\pm$ 0.4116). The furthest embeddings from the English reference are Myanmar (0.9865 $\pm$ 0.7875), Kyrgyz (0.9349 $\pm$ 0.5172), Latin (0.8898 $\pm$ 0.5298).

\null\newpage

\begin{figure*}[ht]
    \centering
    \includegraphics[width=\textwidth]{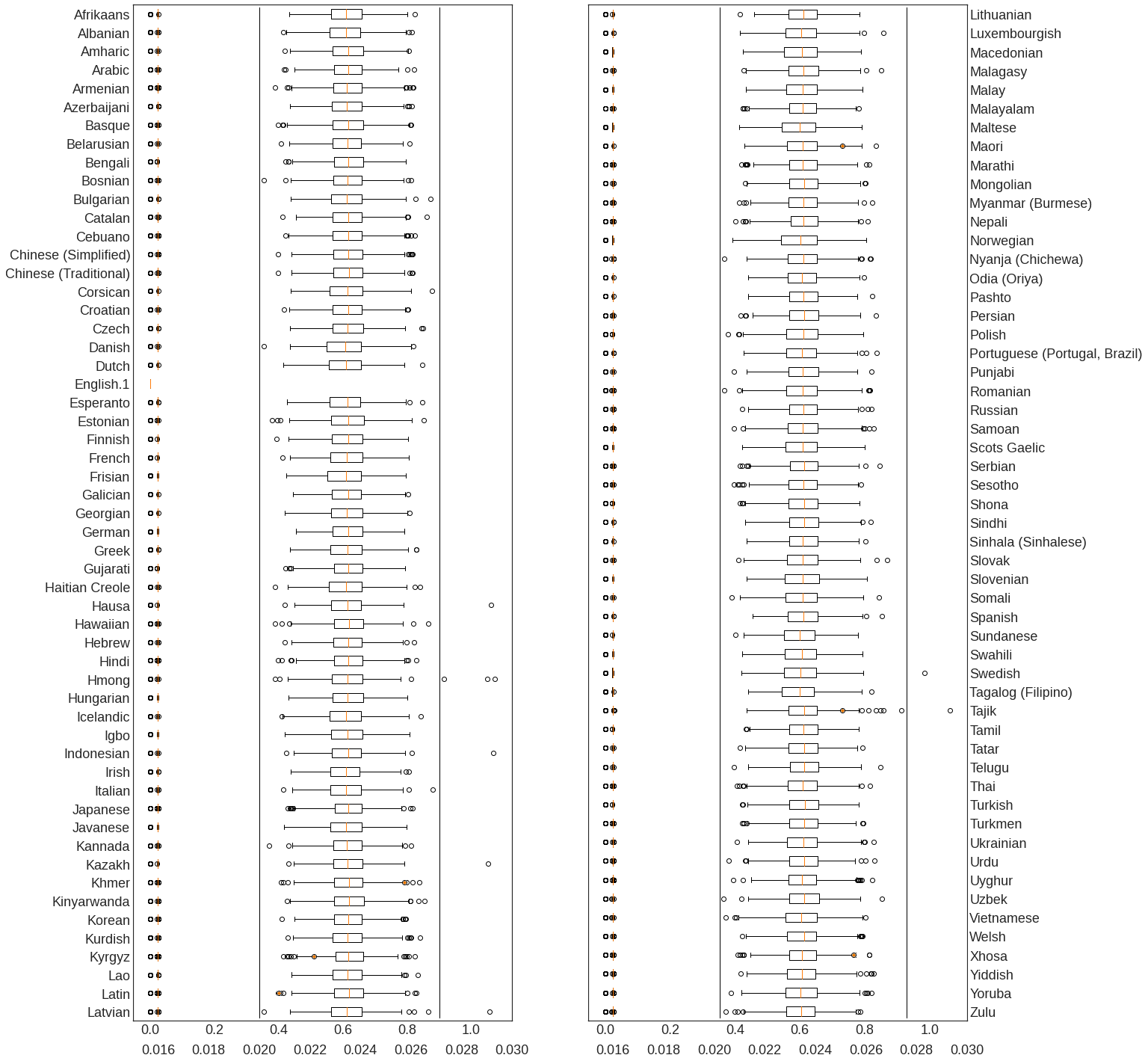}
    \caption{Doc2Vec Euclidean Differences, Full size boxplot with zoomed in plot of the 25-75 percentile on the second x-axis}
    \label{fig:2}
\end{figure*}

\null\newpage

\subsection{Doc2Vec Embeddings}

Boxplot of absolute distance between Doc2Vec embeddings and the English embedding. The closer the language is to English the closer the embeddings are to zero. Doc2Vec embeddings are of 300 dimension and the distance is euclidean. We show the overall graph in addition to the zoomed in portion on the 2nd x-axis.

The closest embeddings to English are much closer than the transformers embeddings regardless of the 300 dimension output. We find closest embeddings are Danish (0.0205 $\pm$ 0.0079), Frisian (0.0206 $\pm$ 0.0078), Tagalog (0.0207 $\pm$ 0.0077). The furthest embeddings from the English reference are Hmong (0.025 $\pm$ 0.0547), Tajik (0.0244 $\pm$ 0.04), Kazakh (0.024 $\pm$ 0.0328).

\begin{figure*}[ht]
    \centering
    \includegraphics[width=\textwidth]{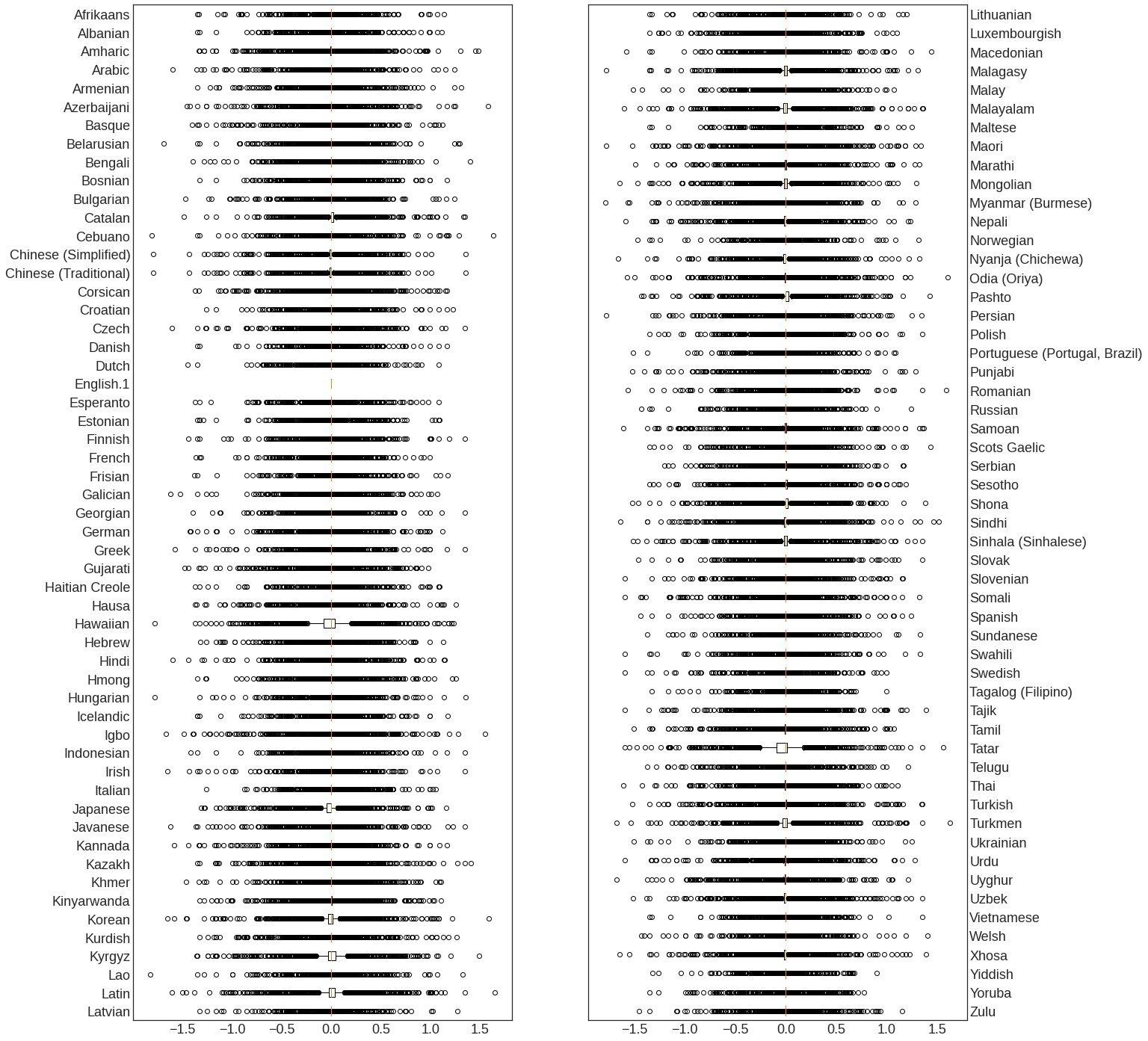}
    \caption{Vader Differences}
    \label{fig:3}
\end{figure*}

\null\newpage

\subsection{Vader Compound Score}

Boxplot of 1D distance between Vader sentiment compound score
Compound score is a 'normalized, weighted composite score' is accurate by summing the valence scores of each word in the lexicon.

The Vader scores with means closest to English are Sesotho (0 $\pm$ 0.2624), Vietnamese (-0.0001 $\pm$ 0.196), Frisian	(0.0002 $\pm$ 0.1850) while the scores further from the reference are Tatar (-0.0388 $\pm$ 0.3227), Japanese (-0.0326 $\pm$ 0.2575), Uyghur (-0.0251 $\pm$ 0.2666).

\null\newpage

\begin{figure*}[ht]
    \centering
    \includegraphics[width=\textwidth]{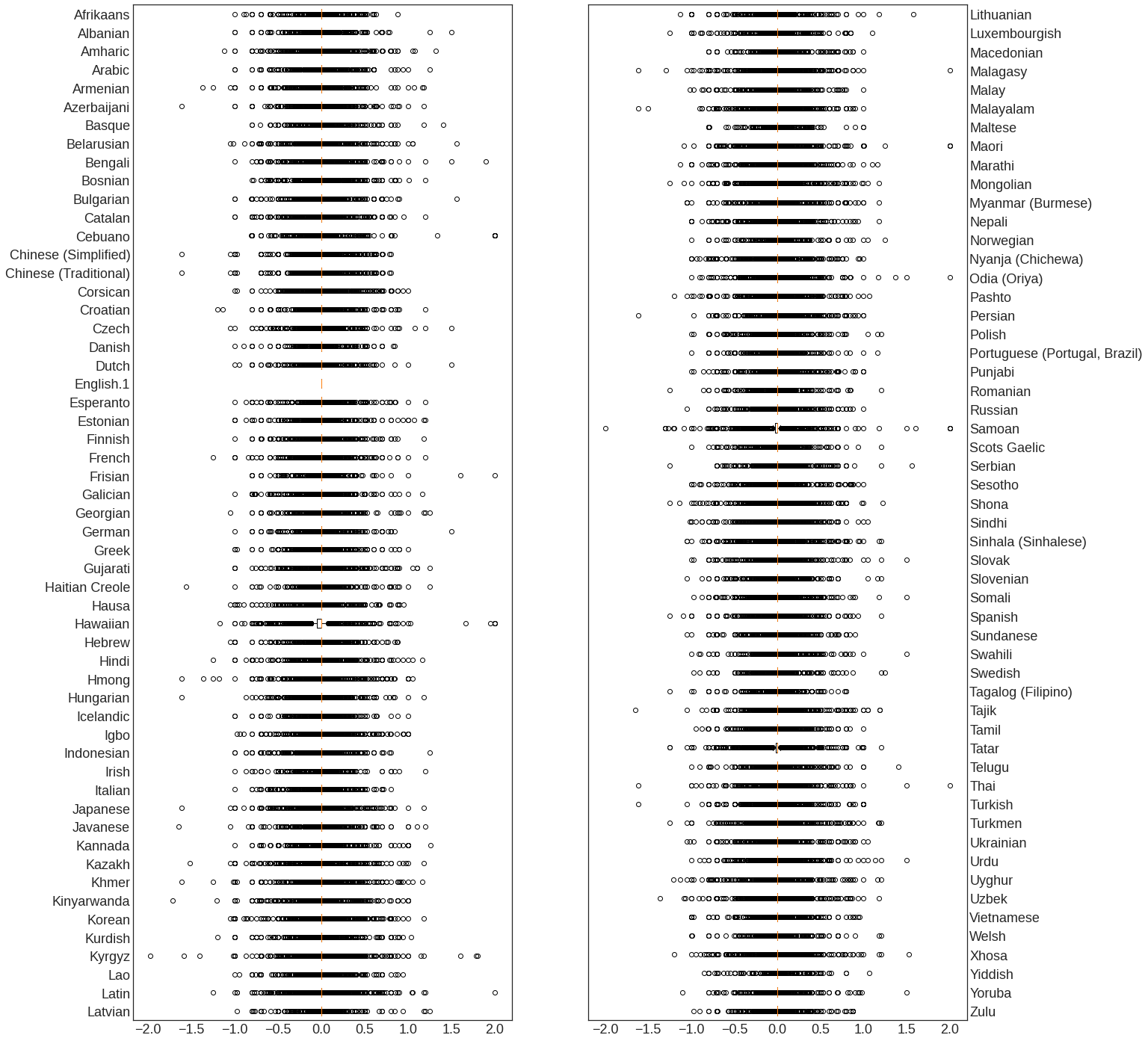}
    \caption{Polarity Differences}
    \label{fig:4}
\end{figure*}

\null\newpage

\subsection{Textblob Polairty Score}
Boxplot of 1D distance between textblob polarity scores
The polarity score is a float within the range [-1.0, 1.0]. The subjectivity is a float within the range [0.0, 1.0] where 0.0 is very objective and 1.0 is very subjective.

The languages with the closest polarity scores are Icelandic (-0.0001 $\pm$ 0.1387), Galician (0.0001 $\pm$ .1738), Odia (Oriya) (0.0001 $\pm$ 0.2172) while the languages furthest from reference polarity are Samoan (-0.0256 $\pm$ 0.2834), Serbian (0.023 $\pm$ 0.1883), Latin (0.0191 $\pm$ 0.2477).

\null\newpage

\begin{figure*}[ht]
    \centering
    \includegraphics[width=\textwidth]{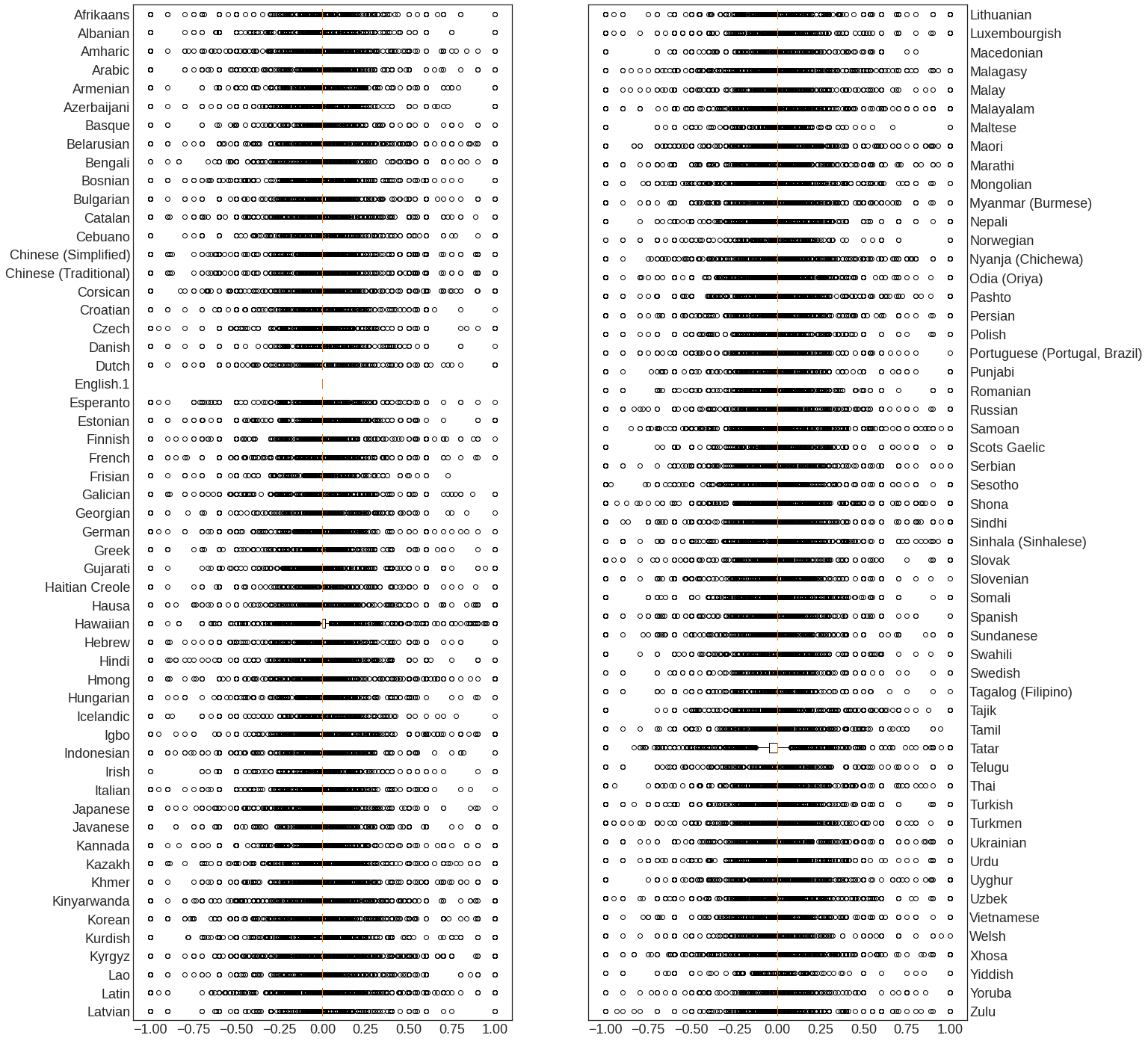}
    \caption{Subjectivity Differences}
    \label{fig:5}
\end{figure*}

\null\newpage

\subsection{Textblob Subjectivity Score}
Boxplot of 1D distance between textblob subjectivity scores
The polarity score is a float within the range [-1.0, 1.0]. The subjectivity is a float within the range [0.0, 1.0] where 0.0 is very objective and 1.0 is very subjective.

The closest embeddings to English are much closer than the BERT embeddings. We find the closest languages are Swahili (0.0000 $\pm$ 0.1767), Swedish (-0.0001 $\pm$ 0.1455), Xhosa (-0.0001 $\pm$ 0.2472) while the languages furthest from the English reference are Tatar (-0.027 $\pm$ 0.2726), Hawaiian (0.0261 $\pm$ 0.2902), Odia (Oriya) (-0.0246 $\pm$ 0.2345).

\null\newpage

\begin{table*}[h]
\centering
\begin{tabular}{||l|c|c|c|c|l|c|c|c|c||}
\hline
\multicolumn{1}{||c|}{\textbf{Language}} &
  \textbf{\begin{tabular}[c]{@{}c@{}}flesch\\ reading\\ ease\end{tabular}} &
  \textbf{\begin{tabular}[c]{@{}c@{}}flesch\\ kincaid\\ grade\end{tabular}} &
  \textbf{\begin{tabular}[c]{@{}c@{}}lexicon\\ count\end{tabular}} &
  \textbf{\begin{tabular}[c]{@{}c@{}}sentence\\ count\end{tabular}} &
  \multicolumn{1}{c|}{\textbf{Language}} &
  \textbf{\begin{tabular}[c]{@{}c@{}}flesch\\ reading\\ ease\end{tabular}} &
  \textbf{\begin{tabular}[c]{@{}c@{}}flesch\\ kincaid\\ grade\end{tabular}} &
  \textbf{\begin{tabular}[c]{@{}c@{}}lexicon\\ count\end{tabular}} &
  \textbf{\begin{tabular}[c]{@{}c@{}}sentence\\ count\end{tabular}} \\ \hline
Afrikaans             & 61.5  & 11.3 & 12786 & 482 & Lithuanian                    & 63.63                      & 10.4 & 12958 & 530 \\ \hline
Albanian              & 62.21 & 11   & 13109 & 508 & Luxembourgish                 & 62.92                      & 10.7 & 12969 & 516 \\ \hline
Amharic               & 65.56 & 9.7  & 13264 & 590 & Macedonian                    & 62.92                      & 10.7 & 12668 & 504 \\ \hline
Arabic                & 67.08 & 9.1  & 13584 & 646 & Malagasy                      & 65.35                      & 9.8  & 12309 & 542 \\ \hline
Armenian              & 61.8  & 11.1 & 12822 & 489 & Malay                         & 67.01                      & 11.2 & 12940 & 440 \\ \hline
Azerbaijani           & 62.61 & 10.8 & 12796 & 503 & Malayalam                     & 65.66                      & 9.7  & 13631 & 609 \\ \hline
Basque                & 62.31 & 11   & 12844 & 500 & Maltese                       & 59.57                      & 12   & 12710 & 448 \\ \hline
Belarusian            & 62.11 & 11   & 13229 & 511 & Maori                         & 69.75                      & 10.2 & 12512 & 469 \\ \hline
Bengali               & 67.59 & 8.9  & 13118 & 641 & Marathi                       & 66.07                      & 9.5  & 12899 & 587 \\ \hline
Bosnian               & 62.31 & 11   & 12819 & 498 & Mongolian                     & 65.25                      & 9.8  & 12648 & 554 \\ \hline
Bulgarian             & 63.32 & 10.6 & 12717 & 514 & Myanmar (Burmese)             & 66.37                      & 9.4  & 12283 & 565 \\ \hline
Catalan               & 63.12 & 10.6 & 12563 & 504 & Nepali                        & 75.54                      & 7.9  & 13068 & 622 \\ \hline
Cebuano               & 61.8  & 11.1 & 12332 & 470 & Norwegian                     & 69.96                      & 10.1 & 13120 & 496 \\ \hline
Chinese (Simplified)  & 66.78 & 9.2  & 12923 & 607 & Nyanja (Chichewa)             & 62.72                      & 10.8 & 12528 & 495 \\ \hline
Chinese (Traditional) & 66.67 & 9.3  & 12863 & 602 & Odia (Oriya)                  & 65.46                      & 9.7  & 13572 & 601 \\ \hline
Corsican              & 63.22 & 10.6 & 12457 & 503 & Pashto                        & 67.18                      & 9.1  & 12529 & 600 \\ \hline
Croatian              & 62.31 & 11   & 12891 & 502 & Persian                       & 68.1                       & 8.7  & 12810 & 642 \\ \hline
Czech                 & 64.54 & 10.1 & 12745 & 543 & Polish                        & 63.12                      & 10.6 & 12944 & 520 \\ \hline
Danish                & 69.45 & 10.3 & 13103 & 486 & Portuguese (Portugal, Brazil) & 62.01                      & 11.1 & 13050 & 502 \\ \hline
Dutch                 & 62.51 & 10.9 & 12721 & 498 & Punjabi                       & 66.67                      & 9.3  & 12442 & 582 \\ \hline
English.1             & 65.29 & 11.9 & 12895 & 414 & Romanian                      & 64.14                      & 10.3 & 12391 & 519 \\ \hline
Esperanto             & 60.18 & 11.8 & 13204 & 475 & Russian                       & 70.57                      & 9.9  & 13226 & 511 \\ \hline
Estonian              & 63.63 & 10.4 & 12856 & 526 & Samoan                        & 60.69                      & 11.6 & 12092 & 443 \\ \hline
Finnish               & 61.6  & 11.2 & 13031 & 493 & Scots Gaelic                  & 63.63                      & 10.4 & 12486 & 511 \\ \hline
French                & 62.21 & 11   & 12708 & 493 & Serbian                       & 64.14                      & 10.3 & 12851 & 537 \\ \hline
Frisian               & 70.06 & 10   & 13097 & 496 & Sesotho                       & 62.51                      & 10.9 & 12218 & 479 \\ \hline
Galician              & 63.22 & 10.6 & 12610 & 509 & Shona                         & 64.34                      & 10.2 & 12518 & 528 \\ \hline
Georgian              & 62.92 & 10.7 & 12676 & 505 & Sindhi                        & 67.28                      & 9    & 12956 & 622 \\ \hline
German                & 69.96 & 10.1 & 12949 & 488 & Sinhala (Sinhalese)           & 67.08                      & 9.1  & 12689 & 603 \\ \hline
Greek                 & 63.83 & 10.4 & 12789 & 528 & Slovak                        & 63.93                      & 10.3 & 12766 & 530 \\ \hline
Gujarati              & 67.59 & 8.9  & 12871 & 629 & Slovenian                     & 63.43                      & 10.5 & 12917 & 525 \\ \hline
Haitian Creole        & 60.89 & 11.5 & 12913 & 477 & Somali                        & 61.9                       & 11.1 & 12569 & 482 \\ \hline
Hausa                 & 61.4  & 11.3 & 12256 & 461 & Spanish                       & 62.92                      & 10.7 & 12906 & 514 \\ \hline
Hawaiian              & 64.95 & 9.9  & 12462 & 539 & Sundanese                     & 60.48                      & 11.7 & 13048 & 475 \\ \hline
Hebrew                & 67.38 & 9    & 12842 & 621 & Swahili                       & 59.47                      & 12   & 12730 & 447 \\ \hline
Hindi                 & 66.78 & 9.2  & 12894 & 604 & Swedish                       & 70.67                      & 9.8  & 12992 & 504 \\ \hline
Hmong                 & 70.87 & 9.7  & 12438 & 485 & Tagalog (Filipino)            & 60.79                      & 11.5 & 12915 & 475 \\ \hline
Hungarian             & 64.54 & 10.1 & 12546 & 534 & Tajik                         & 64.04                      & 10.3 & 12447 & 519 \\ \hline
Icelandic             & 62.51 & 10.9 & 12935 & 507 & Tamil                         & 67.28                      & 9    & 12941 & 621 \\ \hline
Igbo                  & 61.6  & 11.2 & 12267 & 465 & Tatar                         & 60.48                      & 11.7 & 14279 & 520 \\ \hline
Indonesian            & 60.38 & 11.7 & 12953 & 469 & Telugu                        & 74.42                      & 8.4  & 13345 & 603 \\ \hline
Irish                 & 61.4  & 11.3 & 12833 & 483 & Thai                          & 68.13                      & 10.8 & 12755 & 450 \\ \hline
Italian               & 62.82 & 10.8 & 13129 & 521 & Turkish                       & 63.32                      & 10.6 & 12923 & 524 \\ \hline
Japanese              & 73.41 & 8.8  & 14919 & 645 & Turkmen                       & 68.13                      & 10.8 & 13517 & 478 \\ \hline
Javanese              & 59.37 & 12.1 & 12739 & 445 & Ukrainian                     & 63.43                      & 10.5 & 12664 & 515 \\ \hline
Kannada               & 66.67 & 9.3  & 13519 & 633 & Urdu                          & 76.76                      & 7.5  & 12697 & 642 \\ \hline
Kazakh                & 63.22 & 10.6 & 13077 & 527 & Uyghur                        & 65.76                      & 9.6  & 13024 & 585 \\ \hline
Khmer                 & 66.27 & 9.4  & 12714 & 584 & Uzbek                         & 73.21                      & 8.8  & 12528 & 537 \\ \hline
Kinyarwanda           & 67.32 & 11.1 & 13929 & 478 & Vietnamese                    & 61.19                      & 11.4 & 12926 & 482 \\ \hline
Korean                & 67.49 & 9    & 13229 & 641 & Welsh                         & 59.67                      & 12   & 12847 & 454 \\ \hline
Kurdish               & 68.94 & 10.5 & 12892 & 469 & Xhosa                         & 60.99                      & 11.5 & 12532 & 465 \\ \hline
Kyrgyz                & 66.17 & 9.5  & 12232 & 559 & Yiddish                       & 66.67                      & 9.3  & 13333 & 622 \\ \hline
Lao                   & 66.78 & 9.2  & 12315 & 577 & Yoruba                        & 60.79                      & 11.5 & 12766 & 470 \\ \hline
Latin                 & 68.94 & 10.5 & 13874 & 504 & Zulu                          & \multicolumn{1}{l|}{60.38} & 11.7 & 12550 & 454 \\ \hline
Latvian               & 61.9  & 11.1 & 13000 & 498 & \multicolumn{5}{l||}{ } \\ \hline
\end{tabular}
\end{table*}

\null\newpage

\subsection{Text Statistics}
Here we analysis Flesh-reading ease, Flesh-kincaid grade, lexicon count, and sentence count. 

The English Flesch Reading Ease for reference is 65.29. On the top of the list is Urdu (76.76), Nepali (75.54), Telugu (74.42) while on the bottom of the list is Javanese (59.37), Swahili (59.47), Maltese (59.57).

The English Flesch Kincaid Grade for reference is 11.9. On the top of the list is Javanese (12.1), Swahili (12), Maltese (12) while on the bottom fo the list is Urdu (7.5), Nepali (7.9), Telugu (8.4).

The English Lexicon count for reference is 12895. On the top of the list is Japanese (14919), Tatar (14279), Kinyarwanda (13929) while on the bottom of the list is Samoan (12092), Sesotho (12218), Kyrgyz (12232).

The English Sentence count for reference is 414. On the top of the list is Arabic (646), Japanese (645), Urdu (642), while at the bottom of the list is Malay (440), Samoan (443), Javanese (445). 

\null\newpage

\section{Discussion}
\label{sec:Discussion}
As shown the embedding space is quite consistently varied across all 108 languages. Languages with the furthest embeddings (and therefore the best for the most generalized models) are among the languages with the worst translations in the BLEU scores from the Google API. Embedding spaces for large scale transformers have more consistent movement in their feature space while most single metrics (unified sentiment scores) are mostly the same with only outliers not having the same sentiment score. The lexicon nature of those methods do not end up capturing the meaning of context of the sentence and therefore do not end up having their embedding space moved. 

\section{Conclusion}
\label{sec:Conclusion}
Back translation show a significant ability to move the various NLP metrics in many transformer architectures. In this experiment, this text augmentation technique empirically shows back translation acts as a generalizable strategy. Specifically, the lack of good translation allows this technique to move the embedding space in various statistical ways. We show back translation to Tatar moves the embeddings the furthest while translating to Danish would not generalize as well.  

\subsection{Next Steps}
Showing the embedding space is only the first step in deciding which languages are the best for back translation during training. Training transformer architectures on these embeddings is the next step to guarantee a more generalizable model. 

\section*{Acknowledgment}
The authors PeopleTec Technical Fellows program for encouragement and project assistance. The views and conclusions contained in this
paper are those of the authors and should not be interpreted
as representing any funding agencies.

\begin{table*}[]
\centering
\caption{108 Back Translations of a Tweet from the Sentiment 140 Dataset} 
\label{tab:tweet} 
\begin{tabular}{||l|l|l|l||}
\hline
\multicolumn{1}{||c|}{\textbf{Language}} & \multicolumn{1}{c|}{\textbf{Translated Sentence}} & \multicolumn{1}{c|}{\textbf{Language}} & \multicolumn{1}{c||}{\textbf{Translated Sentence}} \\ \hline
Afrikaans & I love Fridays! Held by the pool & Lithuanian & I love Fridays! Laying by the pool \\ \hline
Albanian & I love Fridays! Laying by the pool & Luxembourgish & I love Fridays! Laying of pool \\ \hline
Amharic & I love Friday! Parking at the ready: & Macedonian & Love Friday! Set by the pool \\ \hline
Arabic & I love Fridays! Put on a swimming pool & Malagasy & I love Fridays! Laying by the pool \\ \hline
Armenian & I love Fridays! Laying by the pool & Malay & I love Fridays! Laying by the pool \\ \hline
Azerbaijani & I love Fridays! shoot pool & Malayalam & I love Fridays! laying pool \\ \hline
Basque & I love Fridays! Laying the pool & Maltese & I love Fridays! Placing the pool \\ \hline
Belarusian & I love Fridays! Moore poolside & Maori & I love Friday! Laying in the pit \\ \hline
Bengali & I love Friday! Laying by the pool & Marathi & I love Friday! Laying by the pool \\ \hline
Bosnian & I love Fridays! Laying by the pool & Mongolian & I love Fridays! Laying by the pool \\ \hline
Bulgarian & I love Friday! Laying the pool & Myanmar (Burmese) & I love Friday! Laying the lakes \\ \hline
Catalan & I love Friday! Lie by the pool & Nepali & I love Fridays! Laying by the pool \\ \hline
Cebuano & I love Fridays! Laying the lake & Norwegian & I love Fridays! Laying by the pool \\ \hline
Chinese (Simplified) & I love Fridays! Laying by the pool & Nyanja (Chichewa) & I love the snow! putting pool \\ \hline
Chinese (Traditional) & I love Fridays! Laying by the pool & Odia (Oriya) & I love Fridays! Laying by the pool \\ \hline
Corsican & I love Fridays! Set from pool & Pashto & I love Fridays! Placed by the pool \\ \hline
Croatian & I love Fridays! Laying by the pool & Persian & I'm in love Fridays! Laying by the pool \\ \hline
Czech & I love Fridays! Laying by the pool & Polish & I love Fridays! Lying by the pool \\ \hline
Danish & I love Fridays! Laying by the pool & Portuguese & I love Fridays! Laying by the pool \\ \hline
Dutch & I love Fridays! Laying by the pool & Punjabi & I love Friday! Pool \\ \hline
English & I love Fridays! Laying by the pool & Romanian & I love Fridays! Laying pool \\ \hline
Esperanto & I love Friday! Putting the pool & Russian & I love Fridays! Laying by the pool \\ \hline
Estonian & I love Fridays! laying the pool & Samoan & I love Friday! Laying by the pool \\ \hline
Finnish & I love Fridays! Lay in a pond & Scots Gaelic & I love Friday! Laying with that the swimming \\ \hline
French & I love Fridays! Laying the pool & Serbian & I love Fridays! Laying by the pool \\ \hline
Frisian & I love Fridays! Laying by the pool & Sesotho & I like Friday! Laying by the pool \\ \hline
Galician & I love Friday! Lying in pool & Shona & I love Fridays! putting pool \\ \hline
Georgian & I love Fridays! Laying pool & Sindhi & I love Fridays! Laying Pool \\ \hline
German & I love Fridays! Laying by the pool & Sinhala (Sinhalese) & I love Fridays! Shooting pool \\ \hline
Greek & I love Fridays! Mounting poolside & Slovak & I love Fridays! Laying by the pool \\ \hline
Gujarati & I love Fridays! Laying by the pool & Slovenian & I love heels! Laying by the pool \\ \hline
Haitian Creole & I love Fridays! Laying the pool & Somali & I love Fridays! Laying pool \\ \hline
Hausa & I love Fridays! Laying the pool & Spanish & I love fridays! Lie by the pool \\ \hline
Hawaiian & I love Fridays! Lying by the pool & Sundanese & I love Fridays! Laying by the pool \\ \hline
Hebrew & I love Fridays! And lie by the pool & Swahili & Let me Friday! Keep the pool \\ \hline
Hindi & Mujee like Friday! Laying by the pool & Swedish & I love Fridays! Laying by the pool \\ \hline
Hmong & I love Fridays! Laid by the pool & Tagalog (Filipino) & I love every Friday! Laying by the pool \\ \hline
Hungarian & I love Fridays! Laying the pool & Tajik & I love Fridays! Laying out by the pool \\ \hline
Icelandic & I love Fridays! Laying by the pool & Tamil & I love Fridays! Laying by the pool \\ \hline
Igbo & I love Fridays! To set the pool & Tatar & I am in love on Friday! Laying pool \\ \hline
Indonesian & I love Fridays! Laying by the pool & Telugu & I love Fridays! Laying by the pool \\ \hline
Irish & I love Fridays! To lay by the pool & Thai & I love Fridays! Place the pool \\ \hline
Italian & I love Fridays! Laying in the pool & Turkish & I love Fridays! The pool floor \\ \hline
Japanese & I love Friday! Laying by the pool & Turkmen & I love anna! the foundation stone of the pool \\ \hline
Javanese & I Friday! Laying by the pool & Ukrainian & I love Fridays! Laying pool \\ \hline
Kannada & I love Fridays! Laying by the pool & Urdu & I love Fridays! Laying by the pool \\ \hline
Kazakh & I love Fridays! the construction of a swimming pool & Uyghur & I love Fridays! Pavement with pool \\ \hline
Khmer & I love Fridays! basin & Uzbek & On Friday, I love you! swimming board \\ \hline
Kinyarwanda & I love the fifth! Laying the pool & Vietnamese & I love Friday! Put the pool \\ \hline
Korean & I love Fridays! Lying by the pool & Welsh & I love Fridays! Laying by the pool \\ \hline
Kurdish & I love Fridays! Lying by the pool & Xhosa & I love Fridays! To put it in the pool \\ \hline
Kyrgyz & I love Fridays! by the pool & Yiddish & I love Fridays! Installation by the pool \\ \hline
Lao & I love Friday! By the pool & Yoruba & I love Fridays! Laying by the pool \\ \hline
Latin & I love Fridays? Laying in the pool & Zulu & I love Fridays! Setting up the pool \\ \hline
Latvian & I love Fridays! Putting   the pool & \multicolumn{2}{l||}{} \\ \hline
\end{tabular}
\end{table*}

\begin{table*}[t]
\centering
\caption{Language Metrics Mean and Standard Deviation}
\label{tab:metrics1}
\resizebox{\textwidth}{!}{%
\begin{tabular}{||l|c|c|c|c|c|c|c|c|c||}
\hline
\multicolumn{1}{||c|}{\textbf{Language}} & \textbf{BERT} & \textbf{xlnet} & \textbf{bart} & \textbf{gpt} & \textbf{glove} & \textbf{Doc2Vec} & \textbf{VADER} & \textbf{Polarity} & \textbf{Subjectivity} \\ \hline
Afrikaans & 2.6753 $\pm$ 1.7894 & 36.5315 $\pm$ 20.8224 & 0.7063 $\pm$ 0.4114 & 5.7252 $\pm$ 5.2914 & 0.5019 $\pm$ 0.4145 & 0.0219 $\pm$ 0.0060 & -0.0074 $\pm$ 0.2076 & -0.0008 $\pm$ 0.1559 & -0.0087 $\pm$ 0.1757 \\ \hline
Albanian & 2.7292 $\pm$ 1.8679 & 37.2732 $\pm$ 21.9942 & 0.7189 $\pm$ 0.4193 & 5.9409 $\pm$ 5.3134 & 0.4683 $\pm$ 0.4172 & 0.0216 $\pm$ 0.0064 & -0.0056 $\pm$ 0.2004 & 0.0041 $\pm$ 0.1674 & -0.0163 $\pm$ 0.1759 \\ \hline
Amharic & 4.2694 $\pm$ 1.8715 & 50.2214 $\pm$ 20.3938 & 1.0572 $\pm$ 0.4652 & 10.6803 $\pm$ 6.6559 & 0.6867 $\pm$ 0.4313 & 0.0229 $\pm$ 0.0038 & -0.0137 $\pm$ 0.2816 & 0.0002 $\pm$ 0.2007 & -0.0145 $\pm$ 0.2155 \\ \hline
Arabic & 4.3220 $\pm$ 1.8829 & 47.2316 $\pm$ 19.7599 & 1.0103 $\pm$ 0.4135 & 8.6844 $\pm$ 5.4613 & 0.7693 $\pm$ 0.4834 & 0.0232 $\pm$ 0.0030 & -0.0242 $\pm$ 0.2374 & 0.0035 $\pm$ 0.1754 & -0.0145 $\pm$ 0.1993 \\ \hline
Armenian & 3.3922 $\pm$ 2.0027 & 42.5497 $\pm$ 21.0910 & 0.8890 $\pm$ 0.4417 & 7.5857 $\pm$ 5.9067 & 0.5808 $\pm$ 0.4238 & 0.0224 $\pm$ 0.0051 & -0.0078 $\pm$ 0.2294 & -0.006 $\pm$ 0.1841 & -0.0072 $\pm$ 0.177 \\ \hline
Azerbaijani & 4.5472 $\pm$ 1.8635 & 50.2426 $\pm$ 22.0118 & 1.0690 $\pm$ 0.4685 & 10.3279 $\pm$ 6.4566 & 0.7133 $\pm$ 0.4366 & 0.0229 $\pm$ 0.0039 & -0.0066 $\pm$ 0.2559 & -0.0018 $\pm$ 0.1743 & -0.0081 $\pm$ 0.1693 \\ \hline
Basque & 3.4175 $\pm$ 1.8110 & 43.3499 $\pm$ 22.1301 & 0.8516 $\pm$ 0.4420 & 8.1463 $\pm$ 6.1391 & 0.5432 $\pm$ 0.3888 & 0.0225 $\pm$ 0.0050 & -0.005 $\pm$ 0.2389 & 0.0081 $\pm$ 0.1489 & 0.0008 $\pm$ 0.1524 \\ \hline
Belarusian & 3.4442 $\pm$ 1.8929 & 42.1689 $\pm$ 20.3298 & 0.8668 $\pm$ 0.4231 & 7.4848 $\pm$ 5.4101 & 0.6803 $\pm$ 0.4671 & 0.0225 $\pm$ 0.0049 & -0.005 $\pm$ 0.2368 & 0.0077 $\pm$ 0.2081 & 0.0026 $\pm$ 0.2123 \\ \hline
Bengali & 3.7937 $\pm$ 1.8846 & 48.1835 $\pm$ 21.9692 & 0.9935 $\pm$ 0.4678 & 9.3810 $\pm$ 6.5568 & 0.6196 $\pm$ 0.4292 & 0.0229 $\pm$ 0.0041 & -0.0022 $\pm$ 0.2327 & 0.0116 $\pm$ 0.1953 & 0.0011 $\pm$ 0.1850 \\ \hline
Bosnian & 3.2100 $\pm$ 1.9337 & 40.2262 $\pm$ 21.0845 & 0.8063 $\pm$ 0.4368 & 6.6938 $\pm$ 5.2725 & 0.6114 $\pm$ 0.4419 & 0.0223 $\pm$ 0.0054 & 0.0006 $\pm$ 0.2131 & 0.0112 $\pm$ 0.1806 & -0.0149 $\pm$ 0.1885 \\ \hline
Bulgarian & 3.2032 $\pm$ 1.8526 & 39.5151 $\pm$ 19.3191 & 0.8310 $\pm$ 0.4307 & 6.9228 $\pm$ 5.4705 & 0.6180 $\pm$ 0.4370 & 0.0224 $\pm$ 0.0050 & 0.0031 $\pm$ 0.2178 & 0.0036 $\pm$ 0.1820 & -0.0079 $\pm$ 0.1963 \\ \hline
Catalan & 3.7035 $\pm$ 1.8844 & 44.4201 $\pm$ 21.2358 & 0.9172 $\pm$ 0.4408 & 8.4720 $\pm$ 5.7828 & 0.6943 $\pm$ 0.4918 & 0.0226 $\pm$ 0.0047 & 0.0164 $\pm$ 0.2448 & -0.006 $\pm$ 0.2019 & -0.0137 $\pm$ 0.2155 \\ \hline
Cebuano & 3.1232 $\pm$ 1.9017 & 39.5497 $\pm$ 21.7279 & 0.7828 $\pm$ 0.4377 & 6.9678 $\pm$ 5.8215 & 0.5668 $\pm$ 0.3904 & 0.0221 $\pm$ 0.0057 & 0.0042 $\pm$ 0.2449 & 0.0047 $\pm$ 0.2130 & -0.0005 $\pm$ 0.1686 \\ \hline
Chinese (Simp.) & 3.9368 $\pm$ 1.9524 & 47.6320 $\pm$ 20.6808 & 1.0247 $\pm$ 0.4771 & 8.9017 $\pm$ 5.7305 & 0.7676 $\pm$ 0.4848 & 0.0229 $\pm$ 0.0039 & -0.0166 $\pm$ 0.2556 & 0.0032 $\pm$ 0.1982 & -0.0014 $\pm$ 0.2275 \\ \hline
Chinese (Trad.) & 3.9115 $\pm$ 1.9621 & 47.5426 $\pm$ 20.8729 & 1.0218 $\pm$ 0.4800 & 8.9060 $\pm$ 5.7579 & 0.7681 $\pm$ 0.4932 & 0.0229 $\pm$ 0.0040 & -0.0167 $\pm$ 0.2553 & 0.0031 $\pm$ 0.1986 & -0.0013 $\pm$ 0.2281 \\ \hline
Corsican & 3.3429 $\pm$ 1.9639 & 41.7332 $\pm$ 22.3249 & 0.8559 $\pm$ 0.4762 & 8.1193 $\pm$ 6.3887 & 0.5986 $\pm$ 0.4575 & 0.0219 $\pm$ 0.0060 & -0.002 $\pm$ 0.2604 & 0.0086 $\pm$ 0.1926 & 0.0088 $\pm$ 0.2209 \\ \hline
Croatian & 3.1456 $\pm$ 1.9034 & 40.8590 $\pm$ 21.5791 & 0.7901 $\pm$ 0.4028 & 6.5720 $\pm$ 5.2830 & 0.6174 $\pm$ 0.4898 & 0.0223 $\pm$ 0.0053 & 0.0054 $\pm$ 0.2067 & 0.0105 $\pm$ 0.1721 & -0.0098 $\pm$ 0.1646 \\ \hline
Czech & 3.1310 $\pm$ 1.8350 & 40.1054 $\pm$ 20.7646 & 0.8152 $\pm$ 0.4348 & 6.7120 $\pm$ 5.3675 & 0.5822 $\pm$ 0.4293 & 0.0224 $\pm$ 0.0052 & -0.0055 $\pm$ 0.2321 & -0.0013 $\pm$ 0.1812 & -0.009 $\pm$ 0.1804 \\ \hline
Danish & 2.0758 $\pm$ 1.6713 & 30.1903 $\pm$ 21.0538 & 0.5903 $\pm$ 0.4142 & 4.2798 $\pm$ 4.5897 & 0.4041 $\pm$ 0.4014 & 0.0205 $\pm$ 0.0079 & -0.0011 $\pm$ 0.1806 & 0.0013 $\pm$ 0.1331 & -0.0023 $\pm$ 0.1469 \\ \hline
Dutch & 2.6448 $\pm$ 1.7705 & 35.1354 $\pm$ 21.1426 & 0.6975 $\pm$ 0.4358 & 5.6755 $\pm$ 5.2079 & 0.4924 $\pm$ 0.4191 & 0.0215 $\pm$ 0.0066 & -0.0063 $\pm$ 0.1935 & 0.0009 $\pm$ 0.1618 & -0.0026 $\pm$ 0.1776 \\ \hline
English.1 & 0.0000 $\pm$ 0.0000 & 0.0000 $\pm$ 0.0000 & 0.0000 $\pm$ 0.0000 & 0.0000 $\pm$ 0.0000 & 0.0000 $\pm$ 0.0000 & 0.0000 $\pm$ 0.0000 & 0.0000 $\pm$ 0.0000 & 0.0000 $\pm$ 0.0000 & 0.0000 $\pm$ 0.0000 \\ \hline
Esperanto & 2.3467 $\pm$ 1.7330 & 33.1779 $\pm$ 21.0201 & 0.6306 $\pm$ 0.4066 & 4.8847 $\pm$ 4.8762 & 0.4573 $\pm$ 0.4409 & 0.0212 $\pm$ 0.0071 & -0.0137 $\pm$ 0.2001 & 0.0044 $\pm$ 0.1696 & -0.0198 $\pm$ 0.188 \\ \hline
Estonian & 3.1483 $\pm$ 1.8573 & 41.0236 $\pm$ 21.0431 & 0.8218 $\pm$ 0.4516 & 7.0421 $\pm$ 5.6867 & 0.5903 $\pm$ 0.4497 & 0.0222 $\pm$ 0.0055 & -0.002 $\pm$ 0.208 & 0.0019 $\pm$ 0.1815 & -0.0047 $\pm$ 0.183 \\ \hline
Finnish & 3.0003 $\pm$ 1.8270 & 39.4413 $\pm$ 20.4864 & 0.7904 $\pm$ 0.4394 & 6.4888 $\pm$ 5.2860 & 0.5781 $\pm$ 0.4414 & 0.0222 $\pm$ 0.0055 & 0.0032 $\pm$ 0.2167 & 0.0020 $\pm$ 0.1663 & -0.0027 $\pm$ 0.1833 \\ \hline
French & 3.0026 $\pm$ 1.8019 & 38.7591 $\pm$ 21.2701 & 0.7695 $\pm$ 0.4505 & 6.8370 $\pm$ 5.7311 & 0.5468 $\pm$ 0.4153 & 0.0220 $\pm$ 0.0059 & 0.0023 $\pm$ 0.1985 & -0.0048 $\pm$ 0.1815 & -0.0086 $\pm$ 0.1887 \\ \hline
Frisian & 2.1105 $\pm$ 1.7137 & 31.6068 $\pm$ 21.6522 & 0.5954 $\pm$ 0.4230 & 4.5661 $\pm$ 5.0261 & 0.3772 $\pm$ 0.3800 & 0.0206 $\pm$ 0.0078 & 0.0002 $\pm$ 0.1850 & 0.0003 $\pm$ 0.1522 & -0.0146 $\pm$ 0.1423 \\ \hline
Galician & 3.3084 $\pm$ 1.8283 & 41.5002 $\pm$ 20.6445 & 0.8464 $\pm$ 0.4474 & 7.3423 $\pm$ 5.6460 & 0.6065 $\pm$ 0.4407 & 0.0224 $\pm$ 0.0050 & -0.0054 $\pm$ 0.2207 & 0.0001 $\pm$ 0.1738 & -0.0119 $\pm$ 0.1991 \\ \hline
Georgian & 3.1359 $\pm$ 1.9296 & 41.5282 $\pm$ 21.2911 & 0.8393 $\pm$ 0.4507 & 6.9871 $\pm$ 5.8524 & 0.5619 $\pm$ 0.4255 & 0.0224 $\pm$ 0.0052 & -0.0041 $\pm$ 0.2067 & 0.0068 $\pm$ 0.1653 & -0.0051 $\pm$ 0.1654 \\ \hline
German & 3.0523 $\pm$ 1.8728 & 38.8812 $\pm$ 21.1129 & 0.7681 $\pm$ 0.4358 & 6.8274 $\pm$ 5.7338 & 0.5270 $\pm$ 0.4218 & 0.0222 $\pm$ 0.0056 & -0.0095 $\pm$ 0.2294 & 0.0009 $\pm$ 0.1652 & -0.0149 $\pm$ 0.1882 \\ \hline
Greek & 3.0388 $\pm$ 1.8310 & 38.5410 $\pm$ 21.5254 & 0.7858 $\pm$ 0.4563 & 6.6580 $\pm$ 5.5103 & 0.5591 $\pm$ 0.4279 & 0.0220 $\pm$ 0.0059 & -0.0043 $\pm$ 0.2288 & 0.0015 $\pm$ 0.1600 & -0.0057 $\pm$ 0.1789 \\ \hline
Gujarati & 3.5816 $\pm$ 2.0118 & 45.9618 $\pm$ 21.6391 & 0.9498 $\pm$ 0.4864 & 8.7021 $\pm$ 6.3993 & 0.5882 $\pm$ 0.4797 & 0.0228 $\pm$ 0.0043 & -0.0182 $\pm$ 0.2498 & 0.0122 $\pm$ 0.1941 & 0.0070 $\pm$ 0.1845 \\ \hline
Haitian Creole & 2.4935 $\pm$ 1.7943 & 35.1527 $\pm$ 22.1554 & 0.6506 $\pm$ 0.4146 & 5.4482 $\pm$ 5.2776 & 0.4396 $\pm$ 0.3751 & 0.0213 $\pm$ 0.0070 & -0.0002 $\pm$ 0.2061 & 0.0054 $\pm$ 0.1587 & -0.0066 $\pm$ 0.1645 \\ \hline
Hausa & 3.0370 $\pm$ 1.9173 & 38.3269 $\pm$ 21.4406 & 0.7694 $\pm$ 0.4485 & 6.6423 $\pm$ 5.5684 & 0.5737 $\pm$ 0.4244 & 0.0230 $\pm$ 0.0335 & -0.0156 $\pm$ 0.246 & -0.0056 $\pm$ 0.1898 & 0.0089 $\pm$ 0.2029 \\ \hline
Hawaiian & 4.3917 $\pm$ 2.0020 & 49.2824 $\pm$ 21.7270 & 1.0079 $\pm$ 0.4585 & 10.3520 $\pm$ 6.5060 & 0.8243 $\pm$ 0.4839 & 0.0228 $\pm$ 0.0042 & -0.0236 $\pm$ 0.327 & -0.0188 $\pm$ 0.2746 & 0.0261 $\pm$ 0.2902 \\ \hline
Hebrew & 3.3267 $\pm$ 1.8875 & 43.2773 $\pm$ 19.9098 & 0.8887 $\pm$ 0.4219 & 7.5287 $\pm$ 5.7295 & 0.6314 $\pm$ 0.4726 & 0.0229 $\pm$ 0.0038 & -0.0112 $\pm$ 0.2348 & 0.0037 $\pm$ 0.1807 & -0.0178 $\pm$ 0.1888 \\ \hline
Hindi & 3.7218 $\pm$ 1.8764 & 47.2855 $\pm$ 20.9455 & 0.9400 $\pm$ 0.4265 & 8.8569 $\pm$ 5.9814 & 0.6278 $\pm$ 0.4590 & 0.0230 $\pm$ 0.0038 & -0.0065 $\pm$ 0.2332 & 0.0050 $\pm$ 0.1950 & -0.0081 $\pm$ 0.2007 \\ \hline
Hmong & 3.0967 $\pm$ 1.8916 & 39.2700 $\pm$ 21.5981 & 0.7575 $\pm$ 0.4370 & 6.9259 $\pm$ 5.7573 & 0.5994 $\pm$ 0.5008 & 0.0250 $\pm$ 0.0547 & 0.0145 $\pm$ 0.2360 & 0.0095 $\pm$ 0.2231 & 0.0166 $\pm$ 0.2259 \\ \hline
Hungarian & 3.5887 $\pm$ 1.9075 & 44.1854 $\pm$ 21.7258 & 0.9007 $\pm$ 0.4560 & 8.0363 $\pm$ 5.7908 & 0.6660 $\pm$ 0.4395 & 0.0225 $\pm$ 0.0048 & -0.0005 $\pm$ 0.2233 & -0.0004 $\pm$ 0.1775 & -0.0155 $\pm$ 0.1822 \\ \hline
Icelandic & 2.6405 $\pm$ 1.8053 & 36.4142 $\pm$ 20.9481 & 0.7191 $\pm$ 0.4278 & 5.5187 $\pm$ 5.0761 & 0.4915 $\pm$ 0.4133 & 0.0217 $\pm$ 0.0064 & -0.0015 $\pm$ 0.2054 & -0.0001 $\pm$ 0.1387 & -0.009 $\pm$ 0.1561 \\ \hline
Igbo & 3.4281 $\pm$ 1.9496 & 41.2788 $\pm$ 21.5776 & 0.8291 $\pm$ 0.4534 & 7.7875 $\pm$ 6.0819 & 0.6407 $\pm$ 0.4314 & 0.0221 $\pm$ 0.0057 & -0.0035 $\pm$ 0.2895 & 0.0121 $\pm$ 0.2010 & 0.0225 $\pm$ 0.2272 \\ \hline
Indonesian & 2.9717 $\pm$ 1.8386 & 37.5045 $\pm$ 21.2866 & 0.7086 $\pm$ 0.4175 & 6.3412 $\pm$ 5.3137 & 0.5467 $\pm$ 0.4315 & 0.0230 $\pm$ 0.0337 & -0.0028 $\pm$ 0.2055 & -0.008 $\pm$ 0.1777 & -0.016 $\pm$ 0.1931 \\ \hline
Irish & 2.6636 $\pm$ 1.7391 & 36.4281 $\pm$ 21.8509 & 0.6890 $\pm$ 0.4175 & 6.2791 $\pm$ 5.7358 & 0.4246 $\pm$ 0.3723 & 0.0216 $\pm$ 0.0065 & -0.0106 $\pm$ 0.2102 & -0.0032 $\pm$ 0.1465 & -0.0043 $\pm$ 0.1464 \\ \hline
Italian & 2.8866 $\pm$ 1.8009 & 36.8250 $\pm$ 20.3264 & 0.7365 $\pm$ 0.4360 & 6.3374 $\pm$ 5.4678 & 0.5332 $\pm$ 0.4516 & 0.0217 $\pm$ 0.0063 & 0.0077 $\pm$ 0.1950 & -0.0042 $\pm$ 0.1596 & -0.0183 $\pm$ 0.1867 \\ \hline
Japanese & 4.4044 $\pm$ 1.8878 & 49.7830 $\pm$ 20.5813 & 1.1171 $\pm$ 0.4531 & 10.8755 $\pm$ 6.2951 & 0.7712 $\pm$ 0.4370 & 0.0231 $\pm$ 0.0032 & -0.0326 $\pm$ 0.2575 & -0.0104 $\pm$ 0.2126 & -0.0229 $\pm$ 0.2095 \\ \hline
Javanese & 3.1578 $\pm$ 1.9123 & 39.1957 $\pm$ 21.9393 & 0.7776 $\pm$ 0.4673 & 7.2494 $\pm$ 5.9900 & 0.5517 $\pm$ 0.3895 & 0.0218 $\pm$ 0.0062 & -0.0103 $\pm$ 0.2398 & -0.0003 $\pm$ 0.1801 & -0.0018 $\pm$ 0.1809 \\ \hline
Kannada & 3.9152 $\pm$ 1.8317 & 48.7288 $\pm$ 20.9314 & 1.0231 $\pm$ 0.4551 & 9.9590 $\pm$ 6.5459 & 0.6442 $\pm$ 0.4502 & 0.0229 $\pm$ 0.0038 & -0.0108 $\pm$ 0.2474 & 0.0119 $\pm$ 0.1797 & -0.0068 $\pm$ 0.2071 \\ \hline
Kazakh & 4.1914 $\pm$ 1.8486 & 50.2691 $\pm$ 21.5751 & 1.0625 $\pm$ 0.4802 & 10.6689 $\pm$ 6.6364 & 0.7273 $\pm$ 0.4659 & 0.0240 $\pm$ 0.0328 & -0.0043 $\pm$ 0.2616 & -0.0049 $\pm$ 0.225 & -0.0099 $\pm$ 0.2508 \\ \hline
Khmer & 3.6325 $\pm$ 1.8034 & 45.7669 $\pm$ 19.8676 & 0.9482 $\pm$ 0.4296 & 8.4474 $\pm$ 5.9054 & 0.6743 $\pm$ 0.4798 & 0.0238 $\pm$ 0.0246 & 0.0079 $\pm$ 0.2355 & 0.0110 $\pm$ 0.2183 & 0.0149 $\pm$ 0.2218 \\ \hline
Kinyarwanda & 3.9645 $\pm$ 2.0135 & 45.1762 $\pm$ 21.7999 & 0.9595 $\pm$ 0.4812 & 9.2847 $\pm$ 6.4706 & 0.7150 $\pm$ 0.4545 & 0.0228 $\pm$ 0.0044 & 0.0015 $\pm$ 0.2558 & -0.0035 $\pm$ 0.2228 & -0.0076 $\pm$ 0.2604 \\ \hline
Korean & 4.3485 $\pm$ 1.8834 & 52.2174 $\pm$ 21.6723 & 1.1133 $\pm$ 0.4722 & 11.1366 $\pm$ 6.4485 & 0.7793 $\pm$ 0.4660 & 0.0231 $\pm$ 0.0032 & -0.0074 $\pm$ 0.2906 & 0.0080 $\pm$ 0.2292 & 0.0022 $\pm$ 0.2383 \\ \hline
Kurdish & 3.5824 $\pm$ 2.0262 & 42.2055 $\pm$ 21.8016 & 0.8620 $\pm$ 0.4643 & 8.3174 $\pm$ 6.1876 & 0.6097 $\pm$ 0.4069 & 0.0222 $\pm$ 0.0055 & -0.0124 $\pm$ 0.2565 & -0.0043 $\pm$ 0.1914 & -0.002 $\pm$ 0.2121 \\ \hline
Kyrgyz & 4.7914 $\pm$ 1.9145 & 52.7561 $\pm$ 21.0698 & 1.1718 $\pm$ 0.4801 & 11.6158 $\pm$ 6.4069 & 0.9349 $\pm$ 0.5172 & 0.0236 $\pm$ 0.0158 & -0.0031 $\pm$ 0.3103 & 0.0116 $\pm$ 0.2699 & 0.0123 $\pm$ 0.2555 \\ \hline
Lao & 3.3175 $\pm$ 1.9198 & 43.6135 $\pm$ 21.6109 & 0.8391 $\pm$ 0.4251 & 7.4923 $\pm$ 5.9334 & 0.5750 $\pm$ 0.4177 & 0.0225 $\pm$ 0.0049 & 0.0022 $\pm$ 0.2218 & -0.003 $\pm$ 0.1703 & -0.0022 $\pm$ 0.2022 \\ \hline
Latin & 4.9107 $\pm$ 2.0689 & 52.5410 $\pm$ 21.2892 & 1.2010 $\pm$ 0.5471 & 12.0613 $\pm$ 6.6757 & 0.8898 $\pm$ 0.5298 & 0.0233 $\pm$ 0.0126 & -0.0033 $\pm$ 0.3289 & 0.0191 $\pm$ 0.2477 & -0.0006 $\pm$ 0.2383 \\ \hline
Latvian & 2.6936 $\pm$ 1.7852 & 37.3776 $\pm$ 21.3996 & 0.7204 $\pm$ 0.4138 & 5.6136 $\pm$ 5.0254 & 0.5063 $\pm$ 0.4181 & 0.0228 $\pm$ 0.0334 & -0.0049 $\pm$ 0.1962 & 0.0063 $\pm$ 0.1574 & -0.0083 $\pm$ 0.1664 \\ \hline
\end{tabular}%
}
\end{table*}

\begin{table*}[t]
\centering
\caption{Language Metrics Mean and Standard Deviation}
\label{tab:metrics2}
\resizebox{\textwidth}{!}{%
\begin{tabular}{||l|c|c|c|c|c|c|c|c|c||}
\hline
\multicolumn{1}{||c|}{\textbf{Language}} & \textbf{BERT} & \textbf{xlnet} & \textbf{bart} & \textbf{gpt} & \textbf{glove} & \textbf{Doc2Vec} & \textbf{VADER} & \textbf{Polarity} & \textbf{Subjectivity} \\ \hline
Lithuanian & 3.1533 $\pm$ 1.8887 & 40.9943 $\pm$ 20.4055 & 0.8371 $\pm$ 0.4415 & 6.9824 $\pm$ 5.6158 & 0.5906 $\pm$ 0.4520 & 0.0225 $\pm$ 0.0050 & -0.0073 $\pm$ 0.2184 & 0.0005 $\pm$ 0.1962 & 0.0016 $\pm$ 0.2072 \\ \hline
Luxembourgish & 3.3228 $\pm$ 2.0873 & 41.8774 $\pm$ 22.4604 & 0.8264 $\pm$ 0.4693 & 7.8428 $\pm$ 6.5077 & 0.5650 $\pm$ 0.4214 & 0.0220 $\pm$ 0.0058 & -0.0239 $\pm$ 0.2423 & -0.0054 $\pm$ 0.1864 & -0.016 $\pm$ 0.203 \\ \hline
Macedonian & 2.9089 $\pm$ 1.8076 & 38.2254 $\pm$ 20.8129 & 0.7846 $\pm$ 0.4521 & 6.2868 $\pm$ 5.4331 & 0.5389 $\pm$ 0.4325 & 0.0220 $\pm$ 0.0058 & 0.0033 $\pm$ 0.2172 & 0.0142 $\pm$ 0.1611 & -0.0083 $\pm$ 0.1569 \\ \hline
Malagasy & 3.8505 $\pm$ 1.9621 & 46.8842 $\pm$ 21.4363 & 0.9550 $\pm$ 0.4609 & 8.5786 $\pm$ 5.8147 & 0.7328 $\pm$ 0.4480 & 0.0227 $\pm$ 0.0045 & -0.0096 $\pm$ 0.2748 & -0.0048 $\pm$ 0.2342 & 0.0020 $\pm$ 0.2364 \\ \hline
Malay & 2.9645 $\pm$ 1.7530 & 37.6486 $\pm$ 21.2125 & 0.7135 $\pm$ 0.4255 & 6.6351 $\pm$ 5.4766 & 0.5458 $\pm$ 0.4164 & 0.0221 $\pm$ 0.0057 & 0.0003 $\pm$ 0.2124 & -0.0042 $\pm$ 0.1736 & -0.0039 $\pm$ 0.1938 \\ \hline
Malayalam & 4.8056 $\pm$ 1.7677 & 54.7046 $\pm$ 21.2162 & 1.2327 $\pm$ 0.4586 & 12.2058 $\pm$ 6.3209 & 0.7960 $\pm$ 0.4608 & 0.0233 $\pm$ 0.0022 & -0.0161 $\pm$ 0.3098 & -0.004 $\pm$ 0.2204 & 0.0007 $\pm$ 0.2301 \\ \hline
Maltese & 2.4760 $\pm$ 1.8293 & 33.9896 $\pm$ 22.4921 & 0.6439 $\pm$ 0.4210 & 5.6176 $\pm$ 5.4907 & 0.3895 $\pm$ 0.3353 & 0.0210 $\pm$ 0.0074 & 0.0067 $\pm$ 0.1889 & 0.0027 $\pm$ 0.1226 & -0.0123 $\pm$ 0.1361 \\ \hline
Maori & 3.8948 $\pm$ 1.9485 & 45.9441 $\pm$ 21.5976 & 0.9146 $\pm$ 0.4648 & 9.1761 $\pm$ 6.4592 & 0.6779 $\pm$ 0.4333 & 0.0234 $\pm$ 0.0228 & -0.0074 $\pm$ 0.2998 & -0.0098 $\pm$ 0.2442 & 0.0027 $\pm$ 0.2453 \\ \hline
Marathi & 4.1627 $\pm$ 1.8884 & 50.2687 $\pm$ 21.2404 & 1.0691 $\pm$ 0.4794 & 10.6952 $\pm$ 6.7298 & 0.7018 $\pm$ 0.4285 & 0.0230 $\pm$ 0.0035 & 0.0032 $\pm$ 0.2582 & 0.0026 $\pm$ 0.2057 & -0.0073 $\pm$ 0.227 \\ \hline
Mongolian & 4.2479 $\pm$ 1.9028 & 50.4806 $\pm$ 21.6242 & 1.0513 $\pm$ 0.4389 & 10.5640 $\pm$ 6.3119 & 0.7400 $\pm$ 0.4417 & 0.0229 $\pm$ 0.0040 & -0.0091 $\pm$ 0.2827 & -0.0008 $\pm$ 0.2194 & -0.0101 $\pm$ 0.2155 \\ \hline
Myanmar & 4.7016 $\pm$ 2.1506 & 51.2371 $\pm$ 21.4917 & 1.2359 $\pm$ 0.5540 & 11.9276 $\pm$ 6.6944 & 0.9865 $\pm$ 0.7875 & 0.0229 $\pm$ 0.0040 & -0.0165 $\pm$ 0.2696 & 0.0067 $\pm$ 0.2006 & -0.0082 $\pm$ 0.2115 \\ \hline
Nepali & 4.0612 $\pm$ 1.8466 & 49.6141 $\pm$ 21.5517 & 1.0406 $\pm$ 0.4593 & 10.4111 $\pm$ 6.5645 & 0.6683 $\pm$ 0.4215 & 0.0230 $\pm$ 0.0036 & -0.0231 $\pm$ 0.2732 & -0.0133 $\pm$ 0.1988 & -0.0076 $\pm$ 0.206 \\ \hline
Norwegian & 2.2990 $\pm$ 1.7932 & 31.6301 $\pm$ 20.7609 & 0.6194 $\pm$ 0.4124 & 4.5235 $\pm$ 4.5455 & 0.4440 $\pm$ 0.4034 & 0.0209 $\pm$ 0.0074 & -0.0033 $\pm$ 0.1962 & 0.0058 $\pm$ 0.1506 & -0.0113 $\pm$ 0.1498 \\ \hline
Nyanja & 3.8558 $\pm$ 1.9052 & 45.4260 $\pm$ 20.7191 & 0.9354 $\pm$ 0.4475 & 9.0426 $\pm$ 6.2212 & 0.7501 $\pm$ 0.4549 & 0.0227 $\pm$ 0.0044 & -0.0119 $\pm$ 0.2524 & -0.0116 $\pm$ 0.2206 & 0.0104 $\pm$ 0.2311 \\ \hline
Odia & 4.0704 $\pm$ 1.9244 & 49.8328 $\pm$ 22.4855 & 1.0118 $\pm$ 0.4827 & 10.1309 $\pm$ 6.6468 & 0.6151 $\pm$ 0.3989 & 0.0229 $\pm$ 0.0040 & -0.0105 $\pm$ 0.2847 & 0.0001 $\pm$ 0.2172 & -0.0246 $\pm$ 0.2345 \\ \hline
Pashto & 4.6559 $\pm$ 1.9994 & 51.1143 $\pm$ 21.3486 & 1.0847 $\pm$ 0.4492 & 10.2268 $\pm$ 6.4966 & 0.7500 $\pm$ 0.4439 & 0.0231 $\pm$ 0.0033 & 0.0089 $\pm$ 0.2680 & -0.0038 $\pm$ 0.2037 & -0.0081 $\pm$ 0.2333 \\ \hline
Persian & 4.4312 $\pm$ 1.9276 & 49.8650 $\pm$ 20.9929 & 1.0657 $\pm$ 0.4699 & 9.3400 $\pm$ 5.9010 & 0.7933 $\pm$ 0.5017 & 0.0232 $\pm$ 0.0030 & -0.0071 $\pm$ 0.2667 & 0.0109 $\pm$ 0.2062 & -0.0065 $\pm$ 0.2187 \\ \hline
Polish & 3.3162 $\pm$ 1.8666 & 42.0087 $\pm$ 20.5787 & 0.8431 $\pm$ 0.4436 & 7.3200 $\pm$ 5.5842 & 0.6051 $\pm$ 0.4539 & 0.0225 $\pm$ 0.0049 & 0.0076 $\pm$ 0.2281 & 0.0006 $\pm$ 0.1747 & -0.0077 $\pm$ 0.1833 \\ \hline
Portuguese & 2.8800 $\pm$ 1.8530 & 37.1307 $\pm$ 20.6883 & 0.7394 $\pm$ 0.4346 & 6.2778 $\pm$ 5.4059 & 0.5168 $\pm$ 0.4336 & 0.0219 $\pm$ 0.0060 & -0.003 $\pm$ 0.1948 & 0.0036 $\pm$ 0.1565 & -0.0101 $\pm$ 0.1773 \\ \hline
Punjabi & 3.6739 $\pm$ 1.9488 & 46.4990 $\pm$ 21.4981 & 0.9605 $\pm$ 0.4559 & 8.9806 $\pm$ 6.4451 & 0.6107 $\pm$ 0.4314 & 0.0230 $\pm$ 0.0038 & -0.0054 $\pm$ 0.2478 & 0.0076 $\pm$ 0.1807 & -0.0086 $\pm$ 0.1828 \\ \hline
Romanian & 3.1778 $\pm$ 1.8677 & 40.9882 $\pm$ 21.9662 & 0.8431 $\pm$ 0.4585 & 6.9520 $\pm$ 5.5552 & 0.5621 $\pm$ 0.4292 & 0.0223 $\pm$ 0.0054 & -0.0044 $\pm$ 0.2284 & 0.0015 $\pm$ 0.1592 & -0.0068 $\pm$ 0.1805 \\ \hline
Russian & 3.3236 $\pm$ 1.8926 & 41.1866 $\pm$ 19.9059 & 0.8444 $\pm$ 0.4155 & 7.1935 $\pm$ 5.3579 & 0.6554 $\pm$ 0.4803 & 0.0225 $\pm$ 0.0051 & -0.0062 $\pm$ 0.214 & 0.0073 $\pm$ 0.1843 & -0.0057 $\pm$ 0.2009 \\ \hline
Samoan & 3.8749 $\pm$ 1.9240 & 44.4024 $\pm$ 20.8871 & 0.9006 $\pm$ 0.4179 & 8.9794 $\pm$ 6.2978 & 0.7198 $\pm$ 0.4375 & 0.0225 $\pm$ 0.0049 & -0.0118 $\pm$ 0.2908 & -0.0256 $\pm$ 0.2834 & -0.002 $\pm$ 0.2353 \\ \hline
Scots Gaelic & 3.0683 $\pm$ 1.8886 & 40.3800 $\pm$ 21.6171 & 0.8097 $\pm$ 0.4434 & 6.9198 $\pm$ 5.7215 & 0.5381 $\pm$ 0.3810 & 0.0220 $\pm$ 0.0059 & -0.0002 $\pm$ 0.229 & -0.0067 $\pm$ 0.1535 & -0.0015 $\pm$ 0.178 \\ \hline
Serbian & 3.6279 $\pm$ 1.7905 & 45.0025 $\pm$ 20.4254 & 0.9349 $\pm$ 0.4339 & 8.3136 $\pm$ 5.4780 & 0.7145 $\pm$ 0.5159 & 0.0232 $\pm$ 0.0031 & 0.0112 $\pm$ 0.2407 & 0.0235 $\pm$ 0.1883 & -0.0098 $\pm$ 0.1977 \\ \hline
Sesotho & 3.9322 $\pm$ 1.9724 & 45.9501 $\pm$ 21.0958 & 0.9544 $\pm$ 0.4555 & 8.9511 $\pm$ 5.9904 & 0.7115 $\pm$ 0.4515 & 0.0227 $\pm$ 0.0045 & 0 $\pm$ 0.2624 & -0.0071 $\pm$ 0.2077 & -0.0055 $\pm$ 0.2282 \\ \hline
Shona & 3.8570 $\pm$ 1.8870 & 46.3480 $\pm$ 21.6009 & 0.9486 $\pm$ 0.4390 & 8.8672 $\pm$ 5.9634 & 0.7478 $\pm$ 0.4487 & 0.0227 $\pm$ 0.0045 & -0.0012 $\pm$ 0.2807 & -0.0069 $\pm$ 0.2265 & 0.0103 $\pm$ 0.2553 \\ \hline
Sindhi & 4.8234 $\pm$ 1.9959 & 51.6873 $\pm$ 21.0672 & 1.1107 $\pm$ 0.4948 & 10.9663 $\pm$ 6.5925 & 0.7565 $\pm$ 0.4359 & 0.0232 $\pm$ 0.0029 & -0.0118 $\pm$ 0.2885 & -0.0034 $\pm$ 0.1985 & 0.0059 $\pm$ 0.2285 \\ \hline
Sinhala & 4.4288 $\pm$ 1.8908 & 51.9869 $\pm$ 21.3391 & 1.0986 $\pm$ 0.4804 & 11.4927 $\pm$ 6.7297 & 0.7300 $\pm$ 0.4338 & 0.0230 $\pm$ 0.0037 & -0.0128 $\pm$ 0.2897 & -0.0004 $\pm$ 0.2239 & 0.0072 $\pm$ 0.2366 \\ \hline
Slovak & 3.2220 $\pm$ 1.8268 & 41.3408 $\pm$ 20.6178 & 0.8377 $\pm$ 0.4268 & 7.0250 $\pm$ 5.5250 & 0.6229 $\pm$ 0.4468 & 0.0224 $\pm$ 0.0051 & 0.0038 $\pm$ 0.2269 & -0.0009 $\pm$ 0.1864 & -0.0081 $\pm$ 0.1917 \\ \hline
Slovenian & 3.2271 $\pm$ 1.8310 & 40.4570 $\pm$ 20.0673 & 0.8125 $\pm$ 0.4246 & 6.8675 $\pm$ 5.2968 & 0.6132 $\pm$ 0.4481 & 0.0226 $\pm$ 0.0046 & -0.0011 $\pm$ 0.2357 & -0.0045 $\pm$ 0.1705 & -0.0178 $\pm$ 0.1876 \\ \hline
Somali & 3.6107 $\pm$ 1.9481 & 42.9309 $\pm$ 21.9753 & 0.8895 $\pm$ 0.4700 & 8.5719 $\pm$ 6.2760 & 0.6382 $\pm$ 0.4526 & 0.0222 $\pm$ 0.0056 & -0.0205 $\pm$ 0.2593 & -0.0014 $\pm$ 0.1875 & -0.0087 $\pm$ 0.1975 \\ \hline
Spanish & 3.3843 $\pm$ 1.8808 & 40.6751 $\pm$ 19.7936 & 0.8406 $\pm$ 0.4395 & 7.4043 $\pm$ 5.5196 & 0.6218 $\pm$ 0.4724 & 0.0225 $\pm$ 0.0050 & 0.0054 $\pm$ 0.2251 & -0.0023 $\pm$ 0.1972 & -0.0208 $\pm$ 0.2025 \\ \hline
Sundanese & 2.9151 $\pm$ 1.8701 & 38.0154 $\pm$ 21.1152 & 0.7447 $\pm$ 0.4526 & 6.7278 $\pm$ 5.8818 & 0.4759 $\pm$ 0.3581 & 0.0217 $\pm$ 0.0063 & 0.0014 $\pm$ 0.2095 & 0.0040 $\pm$ 0.1690 & -0.0127 $\pm$ 0.199 \\ \hline
Swahili & 2.8426 $\pm$ 1.8273 & 36.7274 $\pm$ 21.6291 & 0.6900 $\pm$ 0.4117 & 6.2376 $\pm$ 5.5779 & 0.5004 $\pm$ 0.3981 & 0.0217 $\pm$ 0.0064 & 0.0005 $\pm$ 0.2129 & 0.0071 $\pm$ 0.1719 & 0.0000 $\pm$ 0.1767 \\ \hline
Swedish & 2.4565 $\pm$ 1.7945 & 34.4340 $\pm$ 21.1383 & 0.6789 $\pm$ 0.4336 & 5.2523 $\pm$ 5.0355 & 0.4806 $\pm$ 0.4066 & 0.0225 $\pm$ 0.0311 & -0.0136 $\pm$ 0.2135 & 0.0005 $\pm$ 0.1513 & -0.0001 $\pm$ 0.1455 \\ \hline
Tagalog & 2.3192 $\pm$ 1.7408 & 31.9010 $\pm$ 21.4409 & 0.5976 $\pm$ 0.4236 & 5.0015 $\pm$ 4.9439 & 0.4048 $\pm$ 0.3955 & 0.0207 $\pm$ 0.0077 & -0.0007 $\pm$ 0.172 & -0.0025 $\pm$ 0.1454 & -0.0115 $\pm$ 0.1486 \\ \hline
Tajik & 3.6327 $\pm$ 1.8610 & 44.0494 $\pm$ 20.7996 & 0.9278 $\pm$ 0.4599 & 8.6401 $\pm$ 6.3315 & 0.6218 $\pm$ 0.4344 & 0.0244 $\pm$ 0.0400 & 0.0136 $\pm$ 0.2585 & 0.0178 $\pm$ 0.2109 & 0.0062 $\pm$ 0.2101 \\ \hline
Tamil & 4.0502 $\pm$ 2.0863 & 49.8349 $\pm$ 21.9587 & 1.0498 $\pm$ 0.5492 & 10.1832 $\pm$ 6.6707 & 0.6747 $\pm$ 0.5646 & 0.0229 $\pm$ 0.0040 & -0.0168 $\pm$ 0.2599 & -0.0021 $\pm$ 0.1816 & -0.0029 $\pm$ 0.1866 \\ \hline
Tatar & 5.2999 $\pm$ 1.8633 & 55.9961 $\pm$ 21.5646 & 1.3090 $\pm$ 0.5083 & 13.9721 $\pm$ 6.4386 & 0.8622 $\pm$ 0.4289 & 0.0232 $\pm$ 0.0030 & -0.0388 $\pm$ 0.3227 & -0.0181 $\pm$ 0.2454 & -0.027 $\pm$ 0.2726 \\ \hline
Telugu & 3.7709 $\pm$ 1.8097 & 47.7764 $\pm$ 21.5175 & 0.9947 $\pm$ 0.4750 & 9.6866 $\pm$ 6.7028 & 0.5838 $\pm$ 0.3905 & 0.0229 $\pm$ 0.0040 & -0.0151 $\pm$ 0.2313 & 0.0024 $\pm$ 0.1797 & -0.0041 $\pm$ 0.1871 \\ \hline
Thai & 4.2052 $\pm$ 1.9011 & 52.3345 $\pm$ 22.5921 & 1.1586 $\pm$ 0.4787 & 11.6755 $\pm$ 6.6255 & 0.7069 $\pm$ 0.4669 & 0.0231 $\pm$ 0.0032 & -0.0014 $\pm$ 0.2581 & 0.0023 $\pm$ 0.2101 & 0.0036 $\pm$ 0.2037 \\ \hline
Turkish & 4.2634 $\pm$ 1.8235 & 50.1228 $\pm$ 21.2174 & 1.0676 $\pm$ 0.4914 & 10.8648 $\pm$ 6.4401 & 0.7234 $\pm$ 0.4525 & 0.0230 $\pm$ 0.0038 & -0.0038 $\pm$ 0.2788 & 0.0021 $\pm$ 0.2076 & -0.0241 $\pm$ 0.2286 \\ \hline
Turkmen & 4.6553 $\pm$ 1.8657 & 52.6054 $\pm$ 22.3110 & 1.0889 $\pm$ 0.4439 & 11.8237 $\pm$ 6.7022 & 0.8106 $\pm$ 0.4712 & 0.0231 $\pm$ 0.0034 & -0.0115 $\pm$ 0.2965 & 0.0030 $\pm$ 0.2239 & -0.0224 $\pm$ 0.2394 \\ \hline
Ukrainian & 3.5216 $\pm$ 1.9053 & 42.8676 $\pm$ 20.1638 & 0.8892 $\pm$ 0.4196 & 7.7542 $\pm$ 5.6084 & 0.6736 $\pm$ 0.4630 & 0.0226 $\pm$ 0.0047 & 0.0040 $\pm$ 0.2281 & 0.0050 $\pm$ 0.1930 & 0.0020 $\pm$ 0.2158 \\ \hline
Urdu & 4.6288 $\pm$ 1.8598 & 51.8213 $\pm$ 21.1784 & 1.1069 $\pm$ 0.4547 & 10.3984 $\pm$ 6.1689 & 0.7847 $\pm$ 0.4910 & 0.0233 $\pm$ 0.0027 & -0.0079 $\pm$ 0.2553 & 0.0049 $\pm$ 0.2023 & -0.0079 $\pm$ 0.2119 \\ \hline
Uyghur & 4.4782 $\pm$ 2.1602 & 50.8991 $\pm$ 21.7484 & 1.1623 $\pm$ 0.5640 & 11.4469 $\pm$ 6.9056 & 0.7948 $\pm$ 0.6075 & 0.0229 $\pm$ 0.0039 & -0.0251 $\pm$ 0.2666 & -0.0173 $\pm$ 0.2182 & 0.0017 $\pm$ 0.2251 \\ \hline
Uzbek & 4.4620 $\pm$ 1.8179 & 51.5030 $\pm$ 21.9078 & 1.1060 $\pm$ 0.4634 & 11.0946 $\pm$ 6.6179 & 0.8107 $\pm$ 0.4898 & 0.0231 $\pm$ 0.0034 & -0.001 $\pm$ 0.284 & -0.0021 $\pm$ 0.222 & -0.0121 $\pm$ 0.253 \\ \hline
Vietnamese & 3.0241 $\pm$ 1.8411 & 38.3290 $\pm$ 20.3990 & 0.7485 $\pm$ 0.4310 & 6.8053 $\pm$ 5.7475 & 0.5340 $\pm$ 0.4371 & 0.0221 $\pm$ 0.0056 & -0.0001 $\pm$ 0.196 & -0.0021 $\pm$ 0.18 & -0.0063 $\pm$ 0.2022 \\ \hline
Welsh & 2.8045 $\pm$ 1.8026 & 36.3720 $\pm$ 20.9603 & 0.7044 $\pm$ 0.4320 & 6.6622 $\pm$ 5.8721 & 0.4618 $\pm$ 0.3712 & 0.0218 $\pm$ 0.0063 & 0.0021 $\pm$ 0.2133 & 0.0083 $\pm$ 0.1573 & -0.0056 $\pm$ 0.171 \\ \hline
Xhosa & 4.1014 $\pm$ 1.9087 & 48.1886 $\pm$ 21.5661 & 0.9588 $\pm$ 0.4316 & 9.3353 $\pm$ 5.9511 & 0.7828 $\pm$ 0.5315 & 0.0235 $\pm$ 0.0238 & -0.0245 $\pm$ 0.2816 & -0.0022 $\pm$ 0.2329 & -0.0001 $\pm$ 0.2472 \\ \hline
Yiddish & 2.1669 $\pm$ 1.6737 & 37.7814 $\pm$ 19.9013 & 0.7111 $\pm$ 0.3881 & 4.8267 $\pm$ 4.6606 & 0.3952 $\pm$ 0.4116 & 0.0222 $\pm$ 0.0055 & -0.0121 $\pm$ 0.1721 & 0.0025 $\pm$ 0.1317 & -0.0069 $\pm$ 0.1388 \\ \hline
Yoruba & 2.7452 $\pm$ 1.9641 & 34.5736 $\pm$ 21.7557 & 0.6621 $\pm$ 0.4389 & 6.0572 $\pm$ 5.6836 & 0.5101 $\pm$ 0.4320 & 0.0211 $\pm$ 0.0072 & -0.0102 $\pm$ 0.2006 & 0.0027 $\pm$ 0.1647 & 0.0033 $\pm$ 0.1639 \\ \hline
Zulu & 3.0442 $\pm$ 1.9465 & 38.9466 $\pm$ 21.0533 & 0.7747 $\pm$ 0.4438 & 6.7784 $\pm$ 5.7828 & 0.5749 $\pm$ 0.4456 & 0.0220 $\pm$ 0.0059 & -0.0023 $\pm$ 0.2255 & 0.0024 $\pm$ 0.1712 & 0.0030 $\pm$ 0.1826 \\ \hline
\end{tabular}%
}
\end{table*}

\ifCLASSOPTIONcaptionsoff
  \newpage
\fi



%

\bibliographystyle{./bibtex/IEEEtran}
\bibliography{./IEEEexample}
\end{document}